\documentclass{article} 
\usepackage[preprint]{colm2026_conference}

\usepackage{microtype}
\usepackage{hyperref}
\usepackage{url}
\usepackage{booktabs}

\usepackage{graphicx}
\usepackage{subcaption}
\usepackage{amssymb}
\usepackage{mathtools}
\usepackage{amsthm}
\usepackage{algorithm}
\usepackage{algorithmic}
\usepackage{caption}
\usepackage{multirow}
\usepackage{enumitem}

\newcommand{\llada}{LLaDA}
\newcommand{\mdlm}{MDLM}
\newcommand{\ours}{\texttt{S2D2}}
\newcommand{\sdar}{SDAR}
\newcommand{\dream}{Dream}
\newcommand{\bdthree}{BD3}

\usepackage{amsthm}
\theoremstyle{remark}
\newtheorem{remark}{Remark}

\usepackage{lineno}

\definecolor{darkblue}{rgb}{0, 0, 0.5}
\hypersetup{colorlinks=true, citecolor=darkblue, linkcolor=darkblue, urlcolor=darkblue}

\title{\ours{}: Fast Decoding for Diffusion LLMs via Training-Free Self-Speculation}

\author{%
  \vspace{-4mm}\\
  \textbf{Ligong Han\textsuperscript{1,2}\thanks{Correspondence to: Ligong Han (\texttt{lihan@redhat.com})}~~~~Hao Wang\textsuperscript{1,2}~~~~Han Gao\textsuperscript{3}~~~~Kai Xu\textsuperscript{1,2}~~~~Akash Srivastava\textsuperscript{2,4}}\\[0.5mm]
  \textsuperscript{1}Red Hat AI Innovation~~\textsuperscript{2}MIT-IBM Watson AI Lab~~\textsuperscript{3}Iowa State University~~\textsuperscript{4}Core AI, IBM
}

\begin{document}

\ifcolmsubmission
\linenumbers
\fi

\maketitle

\begin{abstract}
Block-diffusion language models offer a promising path toward faster-than-autoregressive generation by combining block-wise autoregressive decoding with within-block parallel denoising. However, in the few-step regime needed for practical acceleration, standard confidence-thresholded decoding is often brittle: aggressive thresholds hurt quality, while conservative thresholds require unnecessary denoising steps. Existing approaches that address this issue either require additional training or incur extra test-time compute.
We present \ours{}, a training-free self-speculative decoding framework for block-diffusion language models. Our key observation is that a block-diffusion model becomes autoregressive when the block size is reduced to one, allowing the same pretrained model to act as both drafter and verifier. \ours{} inserts a speculative verification step into standard block-diffusion decoding and uses lightweight routing policies to decide when verification is worth its cost. This yields a hybrid decoding trajectory in which diffusion proposes tokens in parallel, while the autoregressive mode acts as a local sequence-level critic.
Across three mainstream block-diffusion families, \ours{} consistently improves the accuracy--speed tradeoff over strong confidence-thresholding baselines. On \sdar{}, we observe up to $4.7\times$ speedup over autoregressive decoding, and up to $1.57\times$ over a tuned dynamic decoding baseline while improving accuracy by up to $4.5$ points. On \llada{}2.1-Mini, \ours{} remains complementary to built-in self-correction, including a conservative setting where it is $4.4\times$ faster than the static baseline with slightly higher accuracy.
{Code is available at \url{https://github.com/phymhan/S2D2}.}
\end{abstract}

\section{Introduction}
\label{sec:intro}
Autoregressive (AR) models have driven recent progress in language modeling, especially on reasoning-heavy tasks \
\citep{vaswani2017attention,brown2020language,touvron2023llama,wei2022chain,kojima2022large,jaech2024openai,guo2025deepseek}. Their strict left-to-right generation, however, limits decoding flexibility and inference parallelism. This has motivated diffusion-based language models, which offer a different generation paradigm with potential gains in controllability and speed \
\citep{hoogeboom2021autoregressive,shih2022training,nie2025large,li2022diffusion,schiffsimple,rojas2026improving,labs2025mercury,wang2025diffusion,He_2026_WACV}. Masked diffusion models \
\citep{austin2021structured,sahoo2024simple,shi2024simplified}, first scaled in vision \
\citep{chang2022maskgit,han2022show}, have now been extended to language and shown competitive quality \
\citep{lou2023discrete,gong2024scaling,nie2025large,ye2025dream,rnd1_2025}.

Practical acceleration still requires decoding in only a few denoising steps while preserving efficient Transformer inference (e.g., KV caching). Block diffusion \
\citep{arriola2025block} combines block-wise AR generation (for cache reuse) with within-block diffusion updates (for parallelism), but few-step decoding remains difficult: the common mean-field, token-factorized parameterization \
\citep{xu2024energy,yoo2025redi,zhang2025variational,zhang2026t3d} weakens sequence-level dependencies and can accumulate errors as steps decrease.
Prior work addresses this with explicit sequence-level modeling. EDLM \
\citep{xu2024energy}, for example, introduces an AR energy model and uses self-normalized importance sampling to steer denoising. While effective, this adds training and inference overhead. We instead target a \emph{speed-first} question: can we exploit AR structure at inference time, without extra training, while keeping block-diffusion parallelism?

To study this question, we introduce \ours{}, a \emph{training-free self-speculative decoding} framework for block-diffusion LMs. The key observation is that when block size is reduced to $1$, a block-diffusion model becomes autoregressive and can serve as a verifier. We therefore use standard block-diffusion decoding as the drafter and block-size-$1$ decoding of the same model as the verifier, enabling self-speculation without distillation, auxiliary models, or architectural changes. Since verification adds one extra forward pass, we use lightweight routing policies to invoke it only when worthwhile.

Beyond efficiency, our motivation is also algorithmic. Confidence-threshold decoding can be brittle because acceptance relies on draft confidence alone. Speculative rejection sampling instead uses verifier-normalized acceptance (via the probability ratio), providing a stronger local test for committing drafted tokens. Our goal is not to exactly reproduce block-size-$1$ AR decoding; rather, we use AR verification as a local sequence-level critic inside a hybrid diffusion trajectory.

Empirically, this simple design is often both \emph{faster} and \emph{more accurate} than strong dynamic confidence-threshold baselines. Across three mainstream block-diffusion families, \ours{} improves the accuracy-speed frontier, especially in large-block regimes where standard diffusion decoding is unstable. AR-ness diagnostics further support the view that our verifier provides a stochastic, greedy AR-guided energy correction.
Our contributions are as follows:
\begin{itemize}[nosep,leftmargin=15pt]
    \item We introduce, to the best of our knowledge, the first \emph{training-free self-speculative decoding} method for block-diffusion language models by reusing the block-size-$1$ mode of the same model as a sequence-level verifier.
    \item We develop a practical framework with self-verification masks and lightweight routing policies, enabling plug-and-play acceleration for existing block-diffusion models without additional training.
    \item Through experiments on five models from three major block-diffusion families, we show that \ours{} often improves accuracy while also being faster than competitive dynamic confidence-threshold baselines.
    \item We provide analysis connecting \ours{} to AR-guided residual energy correction, interpreting speculative verification as a stochastic, greedy local preference for lower residual energy.
\end{itemize}

\section{Related Work}
\label{sec:related}

\paragraph{AR-diffusion hybrid language models.}
A key challenge for diffusion LMs is combining parallel token updates with efficient Transformer inference. Block diffusion (BD3)~\citep{arriola2025block} is the first successful AR-diffusion hybrid to combine block-wise AR generation, within-block diffusion decoding, and KV caching, making few-step decoding practical. This design underlies recent block-diffusion LMs such as \llada{} 2.x~\citep{bie2025llada2,bie2026llada2} and \sdar{}~\citep{cheng2025sdar}. Related hybrids include ReFusion~\citep{li2025refusion}, which uses diffusion to plan low-dependency blocks for parallel AR decoding, and Esoteric Language Models~\citep{sahoo2025esotericlanguagemodels}, which combine any-order AR modeling with standard AR decoding. We focus on training-free inference-time acceleration for existing block-diffusion.

\paragraph{Speculative decoding and self-speculation.}
Speculative decoding uses a drafter and verifier with rejection sampling to accelerate generation while preserving the target model distribution~\citep{leviathan2023fast,chen2023accelerating}. In AR models, Draft \& Verify~\citep{zhang2024draft} realizes self-speculation via a weakened version of the same model. In diffusion LMs, ASSD~\citep{guo2025reviving} verifies arbitrary token subsets via any-subset AR modeling~\citep{shih2022training}, but requires specific architectures (e.g., XLNet-style~\citep{yang2019xlnet}) and is not plug-and-play for most pretrained diffusion LMs. BlockSpec~\citep{panblockspec} and SSD~\citep{gao2025self} for diffusion LMs instead uses hierarchical batching over multiple prefix states. Our approach is speed-first and training-free: we reuse the existing block-size-$1$ AR mode of block-diffusion models for single-pass verification. Our routing policies are orthogonal and could be combined with batching-based SSD.

\paragraph{Self-correction and sequence-level correction in diffusion LMs.}
Beyond verifier-based rejection sampling, \llada{}2.1~\citep{bie2026llada2} introduces token editing, which supports an ``unmask early, correct later'' strategy but does not perform verifier-based sequence-level acceptance. Our results show \ours{} is complementary to this mechanism. At a broader level, EDLM~\citep{xu2024energy} and density-ratio-based discrete diffusion methods~\citep{lou2023discrete} also target sequence-level correction, but they rely on additional modeling or extra multi-sample inference. In contrast, we reuse the same pretrained block-diffusion model in AR mode as a local verifier, with no retraining.
Very recently, Introspective DLM~\citep{yu2026introspective} further explores introspective consistency and self-verification from a training-inference co-designed perspective, sharing similar spirits with self-speculation.

\section{Background}
\label{sec:background}

\paragraph{Block-wise autoregressive diffusion decoding.}
We use block-wise autoregressive generation with block size $B$. Given a prompt $x_{1:m}$, decoding proceeds one block at a time: initialize $x^b \leftarrow [\texttt{MASK}]^B$, decode it conditioned on the prompt and finalized blocks, and reuse their KV cache $(\mathbf{K}, \mathbf{V})$ (Algorithm~\ref{alg:sample_bd3}). Let $M_t = \{ i : x_i^b = [\texttt{MASK}] \}$ denote masked positions at diffusion step $t$.

Following masked absorbing-state diffusion, the reverse transition from $t$ to $s < t$ is
\begin{equation}
p_{\theta}(z_s \mid z_t) = q(z_s \mid z_t, x = x_{\theta}(z_t,t)),
\label{eq:subs-posterior}
\end{equation}
where under \mdlm{} \citep{sahoo2024simple} parameterization, for a masked position ($z_t = m$),
\begin{equation}
p_{\theta}(z_s \mid z_t = m) = \mathrm{Cat}\!\left(z_s;\; \frac{1-\alpha_s}{1-\alpha_t} m + \frac{\alpha_s-\alpha_t}{1-\alpha_t} x_{\theta}(z_t,t)\right).
\label{eq:subs-mask}
\end{equation}
Under \mdlm{}, each masked position is independently unmasked with probability $\rho_{t \to s} = \frac{\alpha_s - \alpha_t}{1 - \alpha_t}$ and, if unmasked, assigned a token sampled from $x_{\theta}(z_t,t)$.

In practice, \llada{} \citep{nie2025large} uses few-step confidence-based decoding instead of directly sampling from~\eqref{eq:subs-mask}: a draft pass produces token proposals $\hat{x}$ and confidences $p$ from logits $\ell$, then masked positions are accepted by a fixed schedule or dynamic threshold (Algorithm~\ref{alg:sample_bd3}). Detailed discussion of this transition from \mdlm{} posterior sampling to \llada{} confidence-based decoding is deferred to Appendix~\ref{app:llada_mdlm_decoding}.

\paragraph{Speculative decoding for autoregressive models.}
Speculative decoding (SD) speeds up autoregressive generation by letting a drafter propose multiple tokens and a verifier check them in parallel \citep{leviathan2023fast,chen2023accelerating}. If the draft assigns probability $p_i$ to proposed token $\hat{x}_i$ and the verifier assigns $q_i$ under the target AR distribution, tokens are scanned left-to-right and accepted with probability $\min\!\left(1, \frac{q_i}{p_i}\right)$. At the first rejection, we resample from the residual distribution $(P_{\ell^{\mathrm{ver}}} - P_{\ell})_{+}$.
The procedure preserves the target AR distribution while often accepting multiple tokens per verifier pass.

\begin{figure}[t]
\begin{center}
\includegraphics[width=0.95\linewidth]{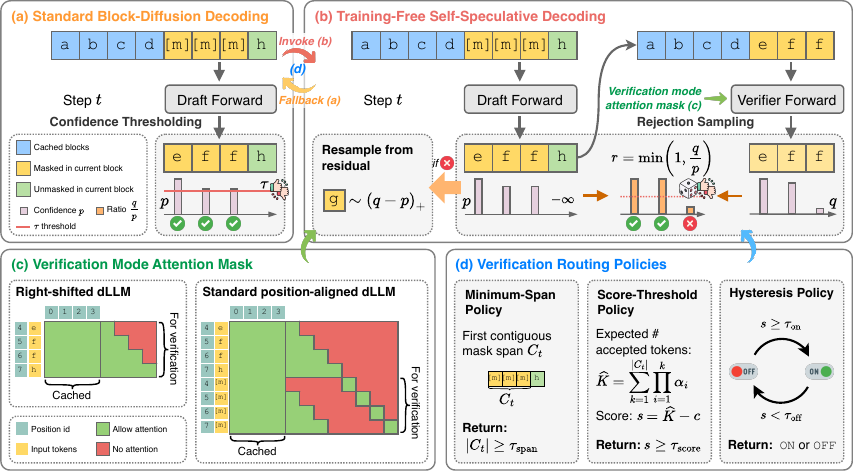}
\end{center}
\caption{Overview of \ours{}. (a) Standard block-diffusion decoding accepts drafted tokens by confidence thresholding. (b) \ours{} inserts a self-speculative verification step: the same model under block-size-$1$ autoregressive masking verifies the first contiguous masked span, accepts tokens by rejection sampling, and falls back to standard diffusion decoding when verification is not invoked or terminates early. (c) Verification-mode attention masks for right-shifted and standard position-aligned diffusion LLMs; we draw the full-block mask for illustration, though in practice only $C_t$ is verified. (d) Lightweight routing policies decide when verification is worth its additional cost.}
\label{fig:method}
\end{figure}

\section{Method}
\label{sec:method}

\subsection{Training-Free Self-Speculative Decoding for Block-Diffusion}
\label{sec:method_overview}

\ours{} reuses a single block-diffusion model in two roles: standard block-diffusion decoding acts as the \emph{drafter}, and the same model with block size $1$ acts as an autoregressive \emph{verifier}. This gives self-speculative decoding without auxiliary models, retraining, or architecture changes (Figure~\ref{fig:method}).
At each denoising step, the drafter proposes tokens $\hat{x}$ with draft probabilities $p$ from logits $\ell$. Instead of immediately applying confidence-threshold acceptance, \ours{} optionally verifies the first contiguous masked span $C_t$: we switch to block-size-$1$ masking, compute verifier probabilities $q$ on drafted tokens in $C_t$, and run standard speculative acceptance left-to-right with probability $\min(1, q_i/p_i)$. At the first rejection, we resample from the residual distribution $(P_{\ell^{\mathrm{ver}}} - P_{\ell})_{+}$ and terminate that speculative segment (Algorithm~\ref{alg:sample_ssd}).

Because verification adds one extra forward pass, it is not always worthwhile (e.g., very short candidate spans). \ours{} therefore uses lightweight \emph{verification routing policies} to decide when to verify and when to fall back to standard confidence-based diffusion decoding. Section~\ref{sec:self_verify_mode} details the verifier mask construction, and Section~\ref{sec:routing_policies} presents routing policies.

\subsection{Self-Verification Mode}
\label{sec:self_verify_mode}
We need verifier probabilities for drafted tokens under a block-size-$1$ autoregressive view, while keeping one shared pretrained block-diffusion model. For a drafted span, the verifier score at position $i$ should condition on drafted tokens to its left and keep position $i$ masked; the key challenge is computing all such scores in parallel.
For position-aligned diffusion LLMs (e.g., \llada{} and \sdar{}), we use the standard ``2$L$ trick'': for a drafted span of length $L$, concatenate the drafted tokens with an all-\texttt{[MASK]} copy at the same positions, and apply
\begin{equation}
\mathbf{M}_{\mathrm{ver}} =
\begin{bmatrix}
\mathbf{A}_L & \mathbf{0}_L \\
\mathbf{A}_L^{<} & \mathbf{I}_L
\end{bmatrix},
\label{eq:ver_mask}
\end{equation}
where $\mathbf{A}_L$ is the causal mask, $\mathbf{A}_L^{<}$ its strict lower-triangular part, and $\mathbf{I}_L$ the identity. This yields all drafted-token verifier confidences in one forward pass (Figure~\ref{fig:method}(c)). For right-shifted models (e.g., \dream{} and Fast-dLLM v2), the standard causal mask already provides the verifier view.

Our setup is related to ASSD \citep{guo2025reviving}, but ASSD requires any-subset AR modeling and dedicated architecture/training; \ours{} is training-free and plug-and-play for existing block-diffusion models, at the cost of verifying only the first contiguous masked span. Even if verification is invoked at every step, \ours{} is still not identical to decoding with block size $1$, since drafting and cache updates need not be fully causal under the original block-diffusion attention pattern. Our optional partially causal drafting variant uses
\begin{equation}
\mathbf{M}_{\mathrm{draft}}^{(j)} =
\begin{bmatrix}
\mathbf{A}_j & \mathbf{0}_{j,\,B-j} \\
\mathbf{1}_{B-j,\,j} & \mathbf{1}_{B-j,\,B-j}
\end{bmatrix},
\label{eq:draft_mask}
\end{equation}
where $j$ is the first masked position in the current block and $x^b_{<j}$ is treated as committed. Here $\mathbf{A}_j\in\{0,1\}^{j\times j}$ is the causal mask on the committed prefix. The four block dimensions are $j\times j$, $j\times(B-j)$, $(B-j)\times j$, and $(B-j)\times(B-j)$, respectively. A visualization is provided in Appendix Figure~\ref{fig:draft_mask}.

\subsection{Verification Routing Policies}
\label{sec:routing_policies}

Verification is useful only when the expected gain from accepting multiple tokens offsets one extra verifier forward pass. We therefore use lightweight routing to decide, at each diffusion step, whether to verify the first contiguous masked span $C_t$ or fall back to confidence-based diffusion decoding (Figure~\ref{fig:method}(d), Algorithm~\ref{alg:verify_policies}).

\paragraph{Expected accepted prefix length.}
Let $C_t=(1,\dots,L)$ after local reindexing. We estimate the expected accepted prefix length as
\begin{equation}
\hat K = \sum_{k=1}^{L} \prod_{i=1}^{k} \alpha_i,
\label{eq:khat}
\end{equation}
where $\alpha_i\in[0,1]$ approximates the acceptance probability at position $i$. We use two proxies: a margin-based form $\alpha_i=\mathbf{1}[m_i\ge\tau_{\text{margin}}]$ (with $m_i$ the top-1 minus top-2 draft probability), and an entropy-based form $\alpha_i=\exp(-\beta\tilde H_i)$ with $\tilde H_i=H_i/\log V$. Additional estimators are deferred to Appendix~\ref{app:estimator_ablation}.

\paragraph{Routing scores and policies.}
We map $\hat K$ to a verification score using either
\begin{equation}
s=\hat K-c\ \text{(static)},\qquad s=\hat K-c\cdot N_{\mathrm{hi}}\ \text{(dynamic)},
\end{equation}
where $c$ is a cost hyperparameter and $N_{\mathrm{hi}}$ counts high-confidence tokens in the current block (i.e., $N_{\mathrm{hi}}=|\{i\in M_t: p_i>\tau\}|$ at diffusion step). We then use one of the following policies:
\begin{itemize}[nosep,leftmargin=15pt]
    \item \textbf{Minimum-span:} invoke verification when $|C_t|\ge\tau_{\mathrm{span}}$. Despite its simplicity, this rule is often surprisingly effective and flexible. For example, setting $\tau_{\mathrm{span}}=1$ always enables verification, setting $\tau_{\mathrm{span}}=B/2$ focuses verification on earlier steps with longer spans, and setting $\tau_{\mathrm{span}}=B$ restricts verification to only the first step of a block.
    \item \textbf{Score-threshold:} invoke verification when $s\ge\tau_{\mathrm{score}}$, using confidence structure rather than span length alone.
    \item \textbf{Hysteresis:} let $h\in\{\textsc{on},\textsc{off}\}$ denote the hysteresis state. If $h=\textsc{on}$ and $s<\tau_{\mathrm{off}}$, set $h\leftarrow\textsc{off}$; if $h=\textsc{off}$ and $s\ge\tau_{\mathrm{on}}$, set $h\leftarrow\textsc{on}$. Verification is invoked iff $h=\textsc{on}$. The motivation is to avoid oscillation between speculative and diffusion modes.
    \item \textbf{Contextual bandit:} we also study a UCB-style contextual bandit router as an additional policy \citep{auer2002finite}; details are deferred to Appendix~\ref{app:bandit_policy}.
\end{itemize}

\subsection{Analysis}
\label{sec:analysis}

We summarize how \ours{} relates to AR-guided residual energy correction. Additional derivations and discussion are deferred to Appendix~\ref{app:energy_connection}.

\paragraph{Not equivalent to global autoregressive decoding.}
\ours{} uses block-size-$1$ AR mode only as a \emph{local verifier}. Verification is applied only on the first contiguous masked span $C_t$, can be skipped by routing, and stops at the first rejection; after that, decoding returns to diffusion with residual resampling. Drafting and KV caching are still generally produced under block-diffusion attention, so the overall trajectory is hybrid rather than globally causal.

\paragraph{Connection to AR-guided residual energy correction.}
EDLM~\citep{xu2024energy} defines a residual energy over diffusion proposals as
\begin{equation}
E_{\phi}(x_0,x_t)
\approx
-\log p_{\mathrm{AR}}(x_0 \mid x_t) + \log p_{\theta}(x_0 \mid x_t) - \log Z.
\label{eq:edlm_energy}
\end{equation}
Both EDLM and \ours{} exploit AR-vs-diffusion discrepancy, but differently: EDLM uses global multi-sample reweighting, while \ours{} applies online local correction through speculative acceptance and residual resampling.

\begin{remark}[Local energy-guided interpretation]
\label{rem:local_energy}
Let $p_i$ be the draft probability of sampled token $\hat{x}_i$ and $q_i$ its verifier probability under block-size-$1$ AR mode. The local residual energy and acceptance form are
\begin{equation}
\label{eq:local_energy}
E_i(\hat{x}_i) := -\log q_i + \log p_i,\qquad
\min\!\left(1,\frac{q_i}{p_i}\right) = \min(1, e^{-E_i(\hat{x}_i)}).
\end{equation}
\label{eq:accept_energy}
Thus, lower-residual-energy drafted tokens are more likely to be accepted, while higher-energy mismatches are corrected via residual resampling.
\end{remark}

This also explains the objective difference: EDLM spends extra test-time compute for quality via global reweighting, whereas \ours{} is designed for acceleration and invokes verification only when expected gain likely amortizes the extra verifier pass.

\section{Experiments}
\label{sec:experiments}

\noindent\textbf{Experimental setup.}
Detailed evaluation setup, including prompt templates and task-specific answer/code extraction, is provided in Appendix~\ref{app:eval_setup}.
\begin{itemize}[leftmargin=0pt,label={},itemsep=1pt,topsep=1pt,parsep=0pt,partopsep=0pt]
\item \textbf{Models.} We evaluate \ours{} on five models from three block-diffusion families: \sdar{}~\citep{cheng2025sdar} (1.7B/4B/8B), Fast-dLLM v2~\citep{wu2025fastdllmtrainingfreeaccelerationdiffusion}, and \llada{}2.1~\citep{bie2026llada2}. \sdar{} and Fast-dLLM v2 are adapted from autoregressive models, whereas \llada{} is trained from scratch. These cover both position-aligned (\sdar{}, \llada{}) and right-shifted (Fast-dLLM v2) architectures.
\item \textbf{Benchmarks.} We report results on \textbf{GSM8K}~\citep{cobbe2021training} (math reasoning), \textbf{MBPP}~\citep{austin2021program} and \textbf{HumanEval}~\citep{chen2021evaluating} (code generation), and \textbf{IFEval}~\citep{zhou2023instruction} (instruction following).
\item \textbf{Decoding baselines.} We compare against standard block-diffusion decoding across block sizes, denoising steps, and static/dynamic confidence schedules. Speedups are reported against the autoregressive baseline (block size $1$), except for Fast-dLLM v2 where we use $\mathrm{B}=4,\ \mathrm{SB}=1$ because $\mathrm{B}=1,\ \mathrm{SB}=1$ is unreliable.
\end{itemize}


\begin{figure}[ht]
    \centering

    \begin{subfigure}[t]{0.25\linewidth}
        \centering
        \includegraphics[width=\linewidth]{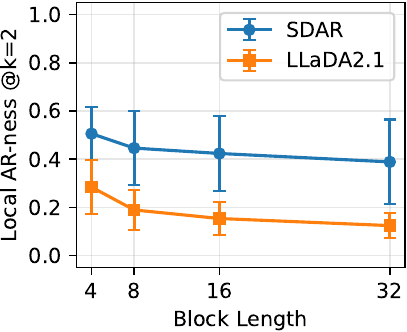}
        \caption{\small \texttt{GSM8K} local AR-ness}
        \label{fig:arness_local_gsm8k}
    \end{subfigure}%
    \begin{subfigure}[t]{0.25\linewidth}
        \centering
        \includegraphics[width=\linewidth]{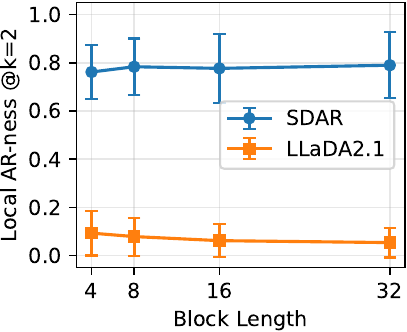}
        \caption{\small \texttt{MBPP} local AR-ness}
        \label{fig:arness_local_mbpp}
    \end{subfigure}%
    \begin{subfigure}[t]{0.25\linewidth}
        \centering
        \includegraphics[width=\linewidth]{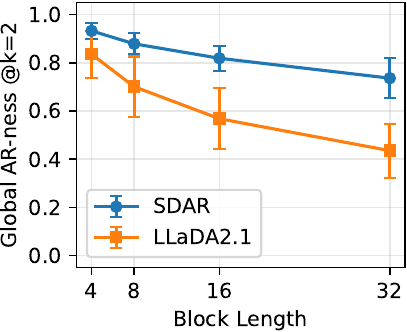}
        \caption{\small \texttt{GSM8K} global AR-ness}
        \label{fig:arness_global_gsm8k}
    \end{subfigure}%
    \begin{subfigure}[t]{0.25\linewidth}
        \centering
        \includegraphics[width=\linewidth]{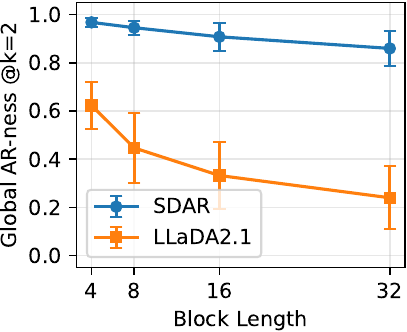}
        \caption{\small \texttt{MBPP} global AR-ness}
        \label{fig:arness_global_mbpp}
    \end{subfigure}

    \vspace{0.2em}

    \begin{subfigure}[t]{0.25\linewidth}
        \centering
        \includegraphics[width=\linewidth]{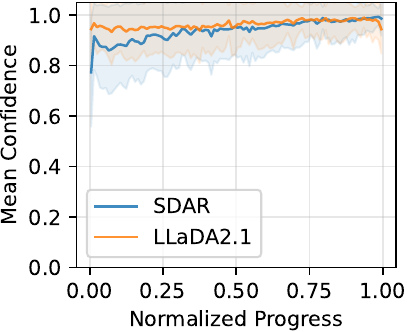}
        \caption{\small \texttt{GSM8K} static conf.}
        \label{fig:conf_static_gsm8k}
    \end{subfigure}%
    \begin{subfigure}[t]{0.25\linewidth}
        \centering
        \includegraphics[width=\linewidth]{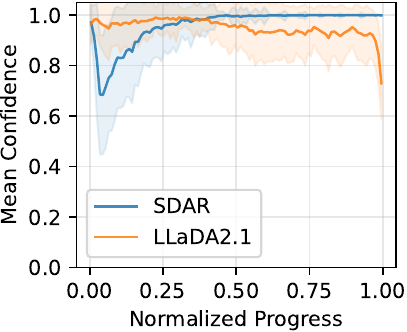}
        \caption{\small \texttt{MBPP} static conf.}
        \label{fig:conf_static_mbpp}
    \end{subfigure}%
    \begin{subfigure}[t]{0.25\linewidth}
        \centering
        \includegraphics[width=\linewidth]{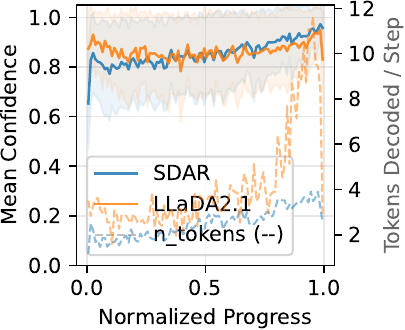}
        \caption{\small \texttt{GSM8K} dynamic conf.}
        \label{fig:conf_dynamic_gsm8k}
    \end{subfigure}%
    \begin{subfigure}[t]{0.25\linewidth}
        \centering
        \includegraphics[width=\linewidth]{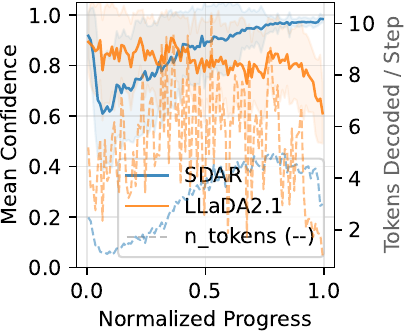}
        \caption{\small \texttt{MBPP} dynamic conf.}
        \label{fig:conf_dynamic_mbpp}
    \end{subfigure}

    \caption{AR-ness (@$k$, $k=2$) and decoding-confidence statistics on GSM8K and MBPP. Top row: local and global AR-ness for \sdar{}-8B-Chat and \llada{} 2.1. Bottom row: normalized decoded-token confidence under static and dynamic diffusion decoding; dashed curves in dynamic decoding indicate the number of decoded tokens per step. Reference accuracies (GSM8K, MBPP): \sdar{}-8B-Chat, AR $(89.3\%, 64.4\%)$ and diffusion $(89.6\%, 61.0\%)$; \llada{} 2.1, AR $(90.8\%, 65.8\%)$ and diffusion $(90.8\%, 67.8\%)$.}
    \label{fig:compare_arness_confidence}
\end{figure}

\begin{table*}[ht]
\centering
\small
\setlength{\tabcolsep}{4pt}
\renewcommand{\arraystretch}{1.15}
\caption{Main accuracy and speedup results on the \sdar{} family. Speedups are measured relative to the autoregressive baseline. For each model, we report standard diffusion decoding at the strongest baseline block size and two representative \ours{} settings: config-A for higher accuracy and config-B for higher speed.}
\label{tab:sdar_main_acc_speedup}
\scalebox{0.9}{
\begin{tabular}{lccccc}
\toprule
\textbf{Method} & {GSM8K} & {MBPP} & {HumanEval} & {IFEval} & {Avg} \\
\midrule
\multicolumn{6}{c}{\textbf{SDAR-1.7B-Chat}} \\
\midrule
AR & 76.3 (1.0$\times$) & 48.4 (1.0$\times$) & 62.2 (1.0$\times$) & 38.6 (1.0$\times$) & 56.4 (1.0$\times$) \\
\midrule
Diffusion (static) & \textbf{77.9} (1.7$\times$) & 13.6 (1.3$\times$) & 51.8 (1.9$\times$) & 43.3 (2.1$\times$) & 46.7 (1.9$\times$) \\
Diffusion (dynamic) & 77.2 (2.7$\times$) & 21.6 (1.3$\times$) & 51.8 (3.4$\times$) & 43.0 (3.8$\times$) & 48.4 (3.1$\times$) \\
\cmidrule(lr){1-6}
\ours{} (config-A) & 77.4 (2.3$\times$) & \textbf{45.4} (1.5$\times$) & 48.8 (2.3$\times$) & \textbf{46.2} (2.7$\times$) & \textbf{54.4} (2.5$\times$) \\
\ours{} (config-B) & 73.8 (\textbf{4.3}$\times$) & 44.4 (\textbf{2.9}$\times$) & \textbf{52.4} (\textbf{4.1}$\times$) & 41.1 (\textbf{5.2}$\times$) & 52.9 (\textbf{4.7}$\times$) \\
\midrule
\multicolumn{6}{c}{\textbf{SDAR-4B-Chat}} \\
\midrule
AR & 88.5 (1.0$\times$) & 59.6 (1.0$\times$) & 75.6 (1.0$\times$) & 54.3 (1.0$\times$) & 69.5 (1.0$\times$) \\
\midrule
Diffusion (static) & \textbf{89.0} (1.4$\times$) & \textbf{57.2} (1.5$\times$) & \textbf{73.8} (1.3$\times$) & 51.0 (1.8$\times$) & 67.7 (1.7$\times$) \\
Diffusion (dynamic) & 88.9 (2.7$\times$) & 55.6 (2.1$\times$) & \textbf{73.8} (1.9$\times$) & 51.7 (3.7$\times$) & 67.5 (3.1$\times$) \\
\cmidrule(lr){1-6}
\ours{} (config-A) & 87.5 (3.7$\times$) & 56.0 (1.3$\times$) & 69.5 (3.1$\times$) & \textbf{59.5} (4.1$\times$) & \textbf{68.1} (3.6$\times$) \\
\ours{} (config-B) & 87.4 (\textbf{4.3}$\times$) & 57.0 (\textbf{2.3}$\times$) & 68.3 (\textbf{3.3}$\times$) & 56.7 (\textbf{5.1}$\times$) & 67.4 (\textbf{4.5}$\times$) \\
\midrule
\multicolumn{6}{c}{\textbf{SDAR-8B-Chat}} \\
\midrule
AR & 89.3 (1.0$\times$) & 64.4 (1.0$\times$) & 75.6 (1.0$\times$) & 57.7 (1.0$\times$) & 71.7 (1.0$\times$) \\
\midrule
Diffusion (static) & \textbf{89.6} (1.4$\times$) & 61.0 (1.4$\times$) & \textbf{79.3} (1.6$\times$) & 52.4 (1.6$\times$) & 70.6 (1.5$\times$) \\
Diffusion (dynamic) & 89.3 (2.6$\times$) & 60.6 (2.1$\times$) & 78.0 (2.9$\times$) & 54.0 (2.7$\times$) & 70.5 (2.6$\times$) \\
\cmidrule(lr){1-6}
\ours{} (config-A) & \textbf{89.6} (2.0$\times$) & \textbf{62.0} (2.1$\times$) & 78.7 (2.2$\times$) & 60.1 (2.2$\times$) & \textbf{72.6} (2.1$\times$) \\
\ours{} (config-B) & 88.3 (\textbf{3.8}$\times$) & 61.4 (\textbf{2.8}$\times$) & 74.4 (\textbf{3.2}$\times$) & \textbf{60.9} (\textbf{3.9}$\times$) & 71.3 (\textbf{3.7}$\times$) \\
\bottomrule
\end{tabular}
}
\end{table*}

\begin{table*}[ht]
\centering
\small
\setlength{\tabcolsep}{4pt}
\renewcommand{\arraystretch}{1.15}
\caption{Main accuracy and speedup results on Fast-dLLM v2. We fix block size $\mathrm{B}=32$ and vary the sub-block size $\mathrm{SB}$. Speedups are measured relative to the autoregressive-style baseline $\mathrm{B}=4, \mathrm{SB}=1$. Config-C corresponds to the normal BD3 without sub-blocks.
}
\label{tab:fast_main_acc_speedup}
\scalebox{0.9}{
\begin{tabular}{lccccc}
\toprule
\textbf{Method} & {GSM8K} & {MBPP} & {HumanEval} & {IFEval} & {Avg} \\
\midrule
AR & 85.1 (1.0$\times$) & 51.8 (1.0$\times$) & 32.9 (1.0$\times$) & 67.7 (1.0$\times$) & 59.4 (1.0$\times$) \\
\midrule
Diffusion (config-A) & \textbf{84.8} (2.5$\times$) & \textbf{50.6} (2.3$\times$) & 14.6 (1.4$\times$) & 63.1 (1.6$\times$) & 53.3 (2.1$\times$) \\
Diffusion (config-B) & 83.2 (3.3$\times$) & 47.8 (2.9$\times$) & 11.0 (1.5$\times$) & 61.6 (1.9$\times$) & 50.9 (2.7$\times$) \\
Diffusion (config-C, SB=32) & 78.3 (3.5$\times$) & 46.2 (3.6$\times$) & 8.5 (1.7$\times$) & 59.4 (1.9$\times$) & 48.1 (2.9$\times$) \\
\cmidrule{1-6}
\ours{} (config-A) & 83.8 (1.6$\times$) & 48.8 (2.0$\times$) & 21.3 (1.5$\times$) & 62.7 (1.3$\times$) & \textbf{54.2} (1.6$\times$) \\
\ours{} (config-B) & 84.2 (3.0$\times$) & 45.0 (3.7$\times$) & \textbf{22.0} (2.0$\times$) & \textbf{64.0} (\textbf{1.9}$\times$) & 53.8 (2.8$\times$) \\
\ours{} (config-C, SB=32) & 82.0 (\textbf{3.9}$\times$) & 45.6 (\textbf{3.8}$\times$) & 19.5 (\textbf{2.5}$\times$) & 63.2 (1.8$\times$) & 52.6 (\textbf{3.1}$\times$) \\
\bottomrule
\end{tabular}
}
\end{table*}

\begin{table*}[ht]
\centering
\small
\setlength{\tabcolsep}{6pt}
\renewcommand{\arraystretch}{1.2}
\caption{Accuracy and speedup results on \llada{}2.1-Mini. Speedups are measured relative to the AR baseline. We compare the built-in editing-based diffusion decoder against \ours{} under two threshold settings. 
}
\label{tab:llada_main_acc_speedup}
\scalebox{0.9}{
\begin{tabular}{lcccccc}
\toprule
& \multicolumn{3}{c}{\(\text{Block-Size}=1\) (AR)} & \multicolumn{3}{c}{\(\tau_{\mathrm{mask}}=1,\ \tau_{\mathrm{edit}}=0.9\)} \\
\cmidrule(lr){2-4}\cmidrule(lr){5-7}
\textbf{Method} & GSM8K & MBPP & Avg & GSM8K & MBPP & Avg \\
\midrule
Diffusion & 90.8 (1.0$\times$) & 65.8 (1.0$\times$) & 78.3 (1.0$\times$) & 90.8 (0.6$\times$) & 67.6 (0.5$\times$) & 79.2 (0.5$\times$) \\
\midrule
& \multicolumn{3}{c}{\(\tau_{\mathrm{mask}}=0.7,\ \tau_{\mathrm{edit}}=0.5\)} & \multicolumn{3}{c}{\(\tau_{\mathrm{mask}}=0.95,\ \tau_{\mathrm{edit}}=0.9\)} \\
\cmidrule(lr){2-4}\cmidrule(lr){5-7}
Diffusion & 89.6 (\textbf{2.7}$\times$) & 57.8 (\textbf{2.8}$\times$) & 73.7 (\textbf{2.7}$\times$) & \textbf{91.0} (1.7$\times$) & 66.4 (1.8$\times$) & 78.7 (1.7$\times$) \\
\ours{} & \textbf{90.8} (2.2$\times$) & \textbf{64.0} (2.1$\times$) & \textbf{77.4} (2.1$\times$) & 89.8 (\textbf{2.1}$\times$) & \textbf{68.8} (\textbf{2.2}$\times$) & \textbf{79.3} (\textbf{2.2}$\times$) \\
\bottomrule
\end{tabular}
}
\end{table*}

\subsection{Main Results}
\label{sec:main_results}

\begin{figure}[t]
\centering
\begin{subfigure}[t]{0.49\linewidth}
    \centering
    \includegraphics[width=0.95\linewidth]{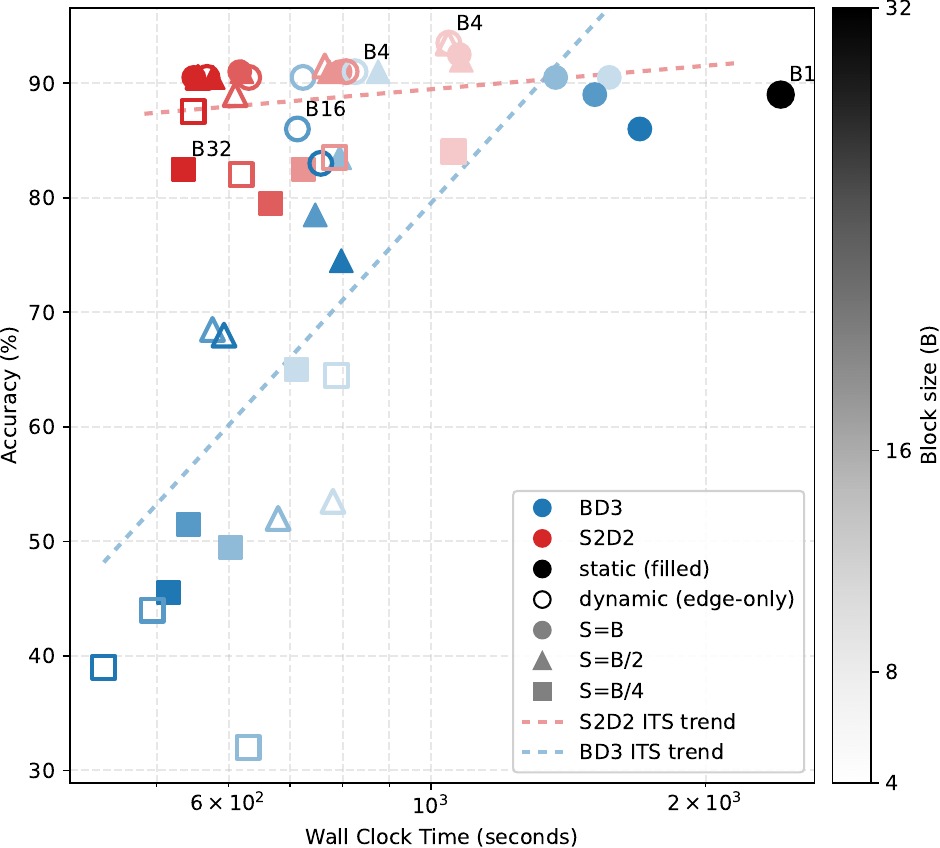}
    \caption{\texttt{GSM8K}}
    \label{fig:sdar8b_gsm8k_acc_speed}
\end{subfigure}\hfill
\begin{subfigure}[t]{0.49\linewidth}
    \centering
    \includegraphics[width=0.95\linewidth]{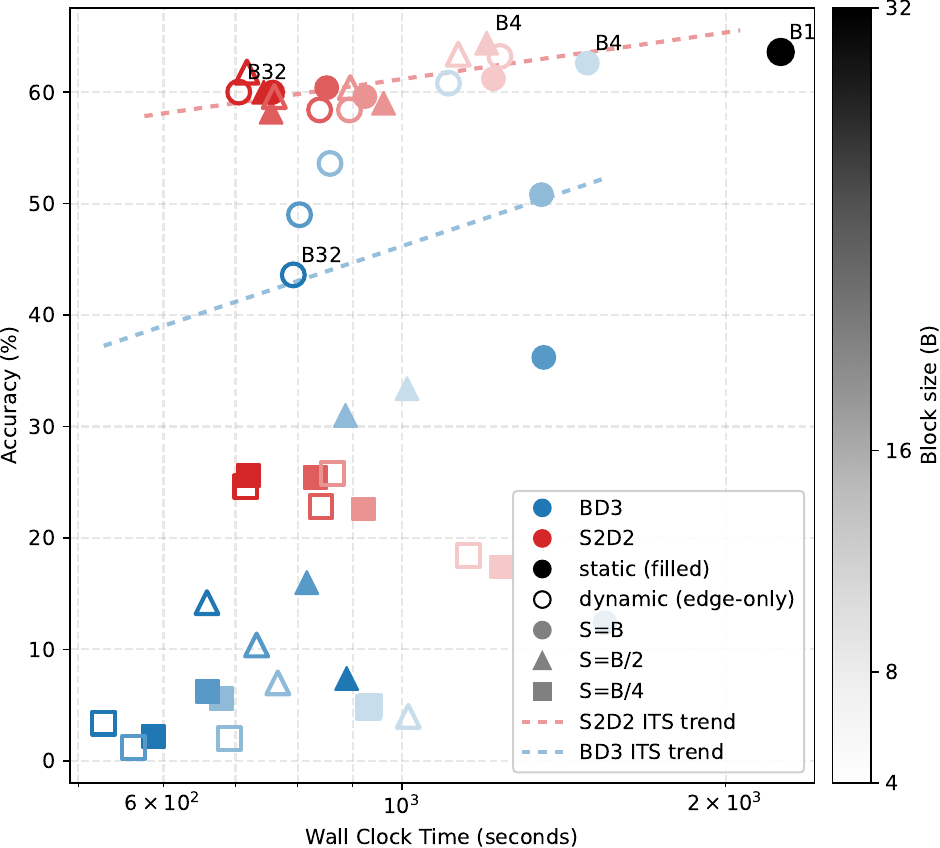}
    \caption{\texttt{MBPP}}
    \label{fig:sdar8b_mbpp_acc_speed}
\end{subfigure}
\caption{Accuracy versus wall-clock time for \sdar{}-8B-Chat on GSM8K and MBPP. ITS denotes inference-time scaling, the ITS trend curve is fitted on points with accuracy $>30\%$. Across block sizes, denoising steps, and decoding schedules, \ours{} generally achieves a better accuracy--speed frontier than \bdthree{}.}
\label{fig:sdar8b_acc_speed}
\end{figure}

\paragraph{Results on \sdar{}.}
Table~\ref{tab:sdar_main_acc_speedup} reports accuracy--speed tradeoffs on \sdar{}-1.7B/4B/8B across GSM8K, MBPP, HumanEval, and IFEval. Here, $B$ denotes block size. Standard \sdar{} decoding is most reliable at small $B$ (especially $\mathrm{B}=4$ or $8$), so we use static/dynamic confidence decoding with $\mathrm{B}=4$ as the diffusion baseline.

For \ours{}, we report two operating points per model: config-A (accuracy-oriented) and config-B (speed-oriented). For \sdar{}-1.7B, config-A uses $\mathrm{B}=4$ with minimum-span routing ($\tau_{\mathrm{span}}=2$), while config-B uses $\mathrm{B}=32$ with always-on speculative verification and AR caching. For \sdar{}-4B, config-A uses $\mathrm{B}=16$ with always-on verification and AR caching, and config-B uses $\mathrm{B}=32$ with the same strategy. For \sdar{}-8B, config-A uses $\mathrm{B}=4$ with score-threshold routing ($\tau_{\mathrm{score}}=0$, static score with $c=1$) and AR caching, while config-B uses $\mathrm{B}=16$ with always-on verification and AR caching.
Across most settings, both configs outperform the dynamic confidence-threshold baseline while remaining faster. Config-A usually gives the best overall accuracy--speed balance, whereas config-B gives larger speedups with a modest accuracy tradeoff. A representative highlight is SDAR-1.7B (config-B): \ours{} reaches 4.7$\times$ speedup over AR decoding, i.e., about 1.57$\times$ over dynamic decoding (4.7/3.1), while improving average accuracy by 4.5 points (52.9 vs. 48.4).

\paragraph{Results on Fast-dLLM v2.}
Table~\ref{tab:fast_main_acc_speedup} reports results on Fast-dLLM v2 with fixed block size $\mathrm{B}=32$ and varying sub-block size $\mathrm{SB}\in\{4,8,16,32\}$. (Here, $B$ is block size and $\mathrm{SB}$ is sub-block size.) The case $\mathrm{SB}=32$ corresponds to standard block-diffusion decoding (config-C). Because Fast-dLLM v2 is unreliable at $\mathrm{B}=1,\mathrm{SB}=1$, we use $\mathrm{B}=4,\mathrm{SB}=1$ as the autoregressive-style baseline.

For the diffusion baseline, we keep the default dynamic-threshold style and fix $\tau=0.9$, with configs A/B/C at $\mathrm{SB}=4/8/32$; sub-block caching is disabled since it is not lossless here. For \ours{}, config-A uses $\mathrm{SB}=4$ with hysteresis routing ($\tau_{\mathrm{on}}=1$, $\tau_{\mathrm{off}}=-5$) and dynamic score ($c=1$), config-B uses the same policy at $\mathrm{SB}=16$, and config-C uses minimum-span routing with $\tau_{\mathrm{span}}=8$ at $\mathrm{SB}=32$.
Compared with the diffusion baseline, \ours{} consistently improves the frontier: config-A improves accuracy with only minor speed loss, config-B improves both accuracy and speed, and config-C recovers much of the large-$\mathrm{SB}$ accuracy drop while still adding speedup. In particular, at config-C (\(\mathrm{SB}=32\)), \ours{} is about 1.07$\times$ faster than dynamic decoding (3.1$\times$ vs. 2.9$\times$) with a +4.5-point average accuracy gain.

\paragraph{Results on \llada{}2.1.}
Table~\ref{tab:llada_main_acc_speedup} reports preliminary GSM8K/MBPP results on \llada{}2.1-Mini~\citep{bie2026llada2}. Unlike standard block-diffusion models, \llada{} supports token editing: previously unmasked tokens can be revised when confidence exceeds $\tau_{\mathrm{edit}}$. This enables an ``unmask early, correct later'' behavior that is related in spirit to self-speculation, but without rejection sampling.

We evaluate two settings: quality mode ($\tau_{\mathrm{mask}}=0.7$, $\tau_{\mathrm{edit}}=0.5$) and a conservative setting ($\tau_{\mathrm{mask}}=0.95$, $\tau_{\mathrm{edit}}=0.9$). In quality mode, \ours{} improves average accuracy over diffusion (77.4\% vs.\ 73.7\%) with moderate speed loss. In the conservative setting, \ours{} improves both accuracy (79.3\% vs.\ 78.7\%) and speedup (2.2$\times$ vs.\ 1.7$\times$), i.e., about 1.3$\times$ faster with +0.6 points. Relative to the static baseline under the same setting (0.5$\times$, 79.2\%), this is 4.4$\times$ faster with slightly higher accuracy (+0.1); note that this static baseline is itself slower than AR (0.5$\times$ vs.\ 1.0$\times$). Overall, \ours{} appears complementary to \llada{}'s built-in self-correction.

\subsection{Analysis and Ablation}
\label{sec:analysis_ablation}
\paragraph{Decoding behavior analysis.}
We analyze baseline diffusion behavior using DiffuCoder local/global AR-ness metrics~\citep{gong2025diffucoder}, where $1$ corresponds to exact left-to-right autoregressive decoding, together with decoded-token confidence; for dynamic decoding, we also report tokens decoded per step (Figure~\ref{fig:compare_arness_confidence}). For \llada{}2.1, we disable editing by setting $\tau_{\mathrm{edit}}=1$. Additional plots are in Appendix Figures~\ref{fig:sdar_arness_confidence} and~\ref{fig:llada_arness_global_mbpp}.

AR-ness shows a task dependence: for \sdar{}, it is higher on MBPP than GSM8K, while for \llada{} 2.1 the trend reverses. This matches the relative competitiveness of AR versus diffusion decoding across tasks. Confidence trajectories also differ: \sdar{} confidence typically rises over decoding, whereas \llada{} often starts high and drops near the end, suggesting stronger AR structure in mathematical reasoning than in code generation.

\paragraph{Inference-time scaling trend.}
Figure~\ref{fig:sdar8b_acc_speed} shows accuracy versus wall-clock time on \sdar{}-8B-Chat across block sizes $B$, denoising steps $S$, schedules, and mask-span settings. \ours{} generally stays on a better frontier and exhibits flatter inference-time scaling than standard diffusion decoding, especially at larger block sizes where the baseline degrades.

\paragraph{Additional ablations.}
Further ablations on token acceptance estimators, routing policies, and rejection-ratio tempering are provided in Appendix~\ref{app:estimator_ablation}, Appendix~\ref{app:routing_ablation}, and Appendix~\ref{app:ratio_tempering}.

\section{Conclusion}

We introduced \ours{}, a training-free self-speculative decoding framework that reuses a single pretrained block-diffusion model in two modes: standard block-diffusion decoding as the drafter and block-size-$1$ autoregressive decoding as the verifier. Across multiple block-diffusion families, this plug-and-play design consistently improves the accuracy-speed tradeoff, often delivering both higher accuracy and lower latency than strong dynamic confidence-thresholding baselines. Our analysis suggests speculative verification acts as a local sequence-level correction mechanism that can be viewed as a stochastic, greedy form of autoregressive energy correction, and we hope this perspective encourages further training-free inference-time methods for diffusion language models.




\newpage
\bibliography{ref}
\bibliographystyle{colm2026_conference}

\newpage
\appendix
\section{Appendix}

\subsection{Algorithm details}
\label{app:algorithm_details}
Algorithms~\ref{alg:block_framework}--\ref{alg:verify_policies} form the full decoding pipeline. Algorithm~\ref{alg:block_framework} is the shared outer loop for block-wise autoregressive decoding, used by both standard diffusion decoding and \ours{}. Within this loop, Algorithm~\ref{alg:sample_bd3} defines dynamic confidence-thresholded diffusion decoding; static decoding is recovered by setting $\tau=1$. Algorithm~\ref{alg:sample_ssd} is our self-speculative block sampler, which adds verifier-based acceptance on top of the same block framework. Algorithm~\ref{alg:verify_policies} specifies the routing procedure \textsc{DoVerify} used by Algorithm~\ref{alg:sample_ssd} to decide when verification is invoked.
\begin{algorithm}[ht]
\caption{Block-wise autoregressive decoding}
\label{alg:block_framework}
\begin{algorithmic}[1]
\REQUIRE Prompt $x_{1:n}$, model $x_\theta$, block size $B$, block sampler $\textsc{SampleBlock}$
\ENSURE Generated sequence $x$

\STATE $x \leftarrow x_{1:n}$, $\mathbf{K}, \mathbf{V} \leftarrow \emptyset$
\WHILE{generation not finished}
    \STATE Initialize a new block $x^b \leftarrow [\texttt{MASK}]^B$
    \STATE $x^b \leftarrow \textsc{SampleBlock}(x_\theta, x^b, \mathbf{K}, \mathbf{V})$
    \STATE $\emptyset, \mathbf{K}^b, \mathbf{V}^b \leftarrow x_\theta(x^b)$
    \STATE $x \leftarrow x \oplus x^b$
    \STATE $(\mathbf{K}, \mathbf{V}) \leftarrow (\mathbf{K} \oplus \mathbf{K}^b,\; \mathbf{V} \oplus \mathbf{V}^b)$
\ENDWHILE
\RETURN $x$
\end{algorithmic}
\end{algorithm}

\begin{algorithm}[ht]
\caption{\textsc{SampleBlock} for standard BD3}
\label{alg:sample_bd3}
\begin{algorithmic}[1]
\REQUIRE Model $x_\theta$, masked block $x^b$, KV cache $(\mathbf{K}, \mathbf{V})$, diffusion steps $T$, confidence threshold $\tau$
\ENSURE Decoded block $x^b$

\FOR{$t = 1$ to $T$}
    \IF{$x^b$ has no \texttt{[MASK]}}
        \STATE \textbf{break}
    \ENDIF
    \STATE $\ell \leftarrow x_\theta(x^b; \mathbf{K}, \mathbf{V})$ \hfill \COMMENT{draft forward}
    \STATE $(\hat{x}, p) \leftarrow \textsc{SampleFromLogits}(\ell)$
    \STATE $M_t \leftarrow \{i : x^b_i = \texttt{[MASK]}\}$
    \STATE $S_t \leftarrow \{i \in M_t : p_i > \tau\}$
    \STATE $S_t \leftarrow S_t \cup \{\arg\max_{i \in M_t} p_i\}$
    \STATE $x^b_{S_t} \leftarrow \hat{x}_{S_t}$
\ENDFOR
\RETURN $x^b$
\end{algorithmic}
\end{algorithm}

\begin{algorithm}[ht]
\caption{\textsc{SampleBlock} for \ours{}}
\label{alg:sample_ssd}
\begin{algorithmic}[1]
\REQUIRE Model $x_\theta$, masked block $x^b$, KV cache $(\mathbf{K}, \mathbf{V})$, diffusion steps $T$, confidence threshold $\tau$
\ENSURE Decoded block $x^b$

\FOR{$t = 1$ to $T$}
    \IF{$x^b$ has no \texttt{[MASK]}}
        \STATE \textbf{break}
    \ENDIF
    \STATE $\ell \leftarrow x_\theta(x^b; \mathbf{K}, \mathbf{V})$ \hfill \COMMENT{draft forward}
    \STATE $(\hat{x}, p) \leftarrow \textsc{SampleFromLogits}(\ell)$
    \STATE $M_t \leftarrow \{i : x^b_i = \texttt{[MASK]}\}$
    \STATE $C_t \leftarrow$ first contiguous mask span in $M_t$
    \IF{\textsc{DoVerify}$(x^b, \ell, p, M_t, C_t)$}
        \STATE $\tilde{x}^b \leftarrow x^b$, \quad $\tilde{x}^b_{C_t} \leftarrow \hat{x}_{C_t}$
        \STATE $\ell^{\mathrm{ver}} \leftarrow x_\theta^{\mathrm{AR}}(\tilde{x}^b; \mathbf{K}, \mathbf{V})$ \hfill \COMMENT{verifier forward}
        \STATE $(\_, q) \leftarrow \textsc{ScoreFromLogits}(\ell^{\mathrm{ver}}, \hat{x}_{C_t})$
        \STATE $S_t \leftarrow \emptyset$
        \FOR{$i \in C_t$ from left to right}
            \STATE Draw $r \sim \mathcal{U}[0,1]$
            \IF{$r < \min(1, q_i / p_i)$}
                \STATE $S_t \leftarrow S_t \cup \{i\}$
            \ELSE
                \STATE Resample $\tilde{x}_i \sim (P_{\ell^{\mathrm{ver}}} - P_{\ell})_{+}$
                \STATE $S_t \leftarrow S_t \cup \{i\}$
                \STATE $\hat{x}_i \leftarrow \tilde{x}_i$
                \STATE \textbf{break}
            \ENDIF
        \ENDFOR
        \STATE $x^b_{S_t} \leftarrow \hat{x}_{S_t}$
    \ELSE
        \STATE $S_t \leftarrow \{i \in M_t : p_i > \tau\}$
        \STATE $S_t \leftarrow S_t \cup \{\arg\max_{i \in M_t} p_i\}$
        \STATE $x^b_{S_t} \leftarrow \hat{x}_{S_t}$
    \ENDIF
\ENDFOR
\RETURN $x^b$
\end{algorithmic}
\end{algorithm}

\begin{algorithm}[ht]
\caption{\textsc{DoVerify}: verification policies}
\label{alg:verify_policies}
\textbf{Require:} Block $x^b$, draft logits $\ell$, draft confidence $p$, masked positions $M_t$, first contiguous mask span $C_t$, thresholds $\tau_{\mathrm{span}}, \tau_{\mathrm{score}}, \tau_{\mathrm{on}}, \tau_{\mathrm{off}}$, hysteresis state $h$\\
\textbf{Ensure:} Boolean decision $d \in \{\texttt{True}, \texttt{False}\}$

\vspace{0.25em}
\textbf{Minimum-Span Policy}
\begin{algorithmic}[1]
\STATE \textbf{return} $(|C_t| \ge \tau_{\mathrm{span}})$
\end{algorithmic}

\vspace{0.25em}
\textbf{Score-Threshold Policy}
\begin{algorithmic}[1]
\STATE $s \leftarrow \textsc{ComputeVerifyScore}(x^b, \ell, p, M_t, C_t)$
\STATE \textbf{return} $(s \ge \tau_{\mathrm{score}})$
\end{algorithmic}

\vspace{0.25em}
\textbf{Hysteresis Policy}
\begin{algorithmic}[1]
\STATE $s \leftarrow \textsc{ComputeVerifyScore}(x^b, \ell, p, M_t, C_t)$
\IF{$h=\texttt{ON}$ and $s<\tau_{\mathrm{off}}$}
    \STATE $h \leftarrow \texttt{OFF}$
\ELSIF{$h=\texttt{OFF}$ and $s\geq\tau_{\mathrm{on}}$}
    \STATE $h \leftarrow \texttt{ON}$
\ENDIF
\STATE \textbf{return} $(h=\texttt{ON})$
\end{algorithmic}
\end{algorithm}

\begin{figure}[ht]
\begin{center}
\includegraphics[width=1\linewidth]{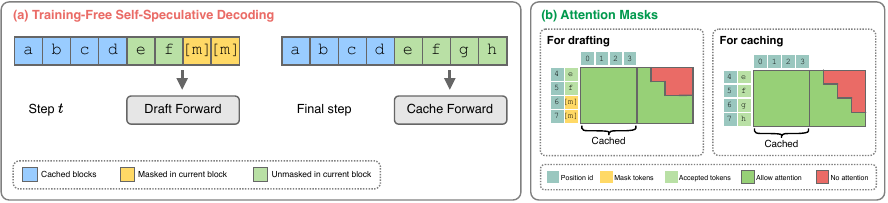}
\end{center}
\caption{Drafting and caching with autoregressive attention masks.}
\label{fig:draft_mask}
\end{figure}

\paragraph{Optional AR-like drafting/caching mask.}
Figure~\ref{fig:draft_mask} visualizes the optional partially causal drafting/caching mask used to make the trajectory more AR-like. The mask definition is given in Eq.~\ref{eq:draft_mask} in Section~\ref{sec:self_verify_mode}.

\subsection{From \mdlm{} posterior sampling to \llada{} confidence-based decoding}
\label{app:llada_mdlm_decoding}

For absorbing-state discrete diffusion, the reverse transition under the SUBS parameterization of \mdlm{} \citep{sahoo2024simple} is
\begin{equation}
p_{\theta}(z_s \mid z_t) = q(z_s \mid z_t, x = x_{\theta}(z_t,t)).
\end{equation}
For a masked position $z_t = m$, Eq.~\eqref{eq:subs-mask} can be equivalently interpreted as a two-stage procedure: first, sample whether the position remains masked or becomes visible with probability
\begin{equation}
\rho_{t\to s} = \frac{\alpha_s-\alpha_t}{1-\alpha_t},
\end{equation}
and second, if the position is unmasked, sample its token from the model prediction $x_{\theta}(z_t,t)$. Thus, exact reverse sampling factorizes across masked positions.

In contrast, \llada{} decoding \citep{nie2025large} typically operates in a few-step regime and does not directly sample from this factorized posterior. Instead, after a draft forward pass, it selects which positions to unmask using token confidence, either according to a fixed schedule or a dynamic threshold. This can be viewed as replacing independent Bernoulli reveal decisions with a confidence-prioritized selection rule.

We find this interpretation useful for understanding why confidence-based decoding often works well in practice. First, the exact reverse posterior in Eq.~\eqref{eq:subs-mask} is conditionally factorized across positions, which introduces a \textbf{mean-field} approximation at the block level. In the few-step regime, confidence-based selection partially mitigates this approximation by introducing dependence among positions through competitive reveal decisions: instead of revealing positions independently, the decoder preferentially commits to the most reliable tokens first. Second, posterior sampling in diffusion can be viewed as approximating an intractable integral using Monte Carlo samples, as discussed in \textbf{diffusion posterior sampling (DPS)} \citep{chung2022diffusion}. From this perspective, practical decoding effectively uses a single sample to approximate the reverse update, and choosing the highest-confidence proposals can reduce the approximation error of this one-sample estimator. Dynamic confidence thresholding strengthens this effect by adaptively revealing as many positions as possible once their confidence exceeds a preset threshold, yielding substantial acceleration while usually preserving generation quality.

\subsection{Connection to residual energy}
\label{app:energy_connection}

We further clarify the connection between \ours{} and AR-guided residual energy correction.

\paragraph{Residual energy and speculative acceptance.}
EDLM defines a residual energy over diffusion proposals of the form
\begin{align}
E_{\phi}(x_0, x_t)
&\approx -\log q(x_0 \mid x_t) + \log p_{\theta}(x_0 \mid x_t) - \log Z \nonumber\\
&= -\log q(x_0) + \log p_{\theta}(x_0 \mid x_t) + \bigl(\log q(\bar x_0) - \log Z\bigr),
\label{eq:app_edlm_energy}
\end{align}
where $q(x_0)$ denotes the autoregressive verification density and $p_{\theta}(x_0 \mid x_t)$ is the diffusion proposal distribution \citep{xu2024energy}. Ignoring the additive constant $\log q(\bar x_0) - \log Z$, (where $\bar x_0$ is the unmasked tokens in $x_t$) this is exactly the negative log density ratio:
\begin{equation}
E_{\phi}(x_0, x_t)
=
-\log \frac{q(x_0)}{p_{\theta}(x_0 \mid x_t)}
\quad \text{up to an additive constant.}
\label{eq:app_energy_ratio}
\end{equation}
In our setting, for a drafted token $\hat{x}_i$, let $p_i$ denote the draft probability under the diffusion forward pass and let $q_i$ denote the verifier probability under the block-size-$1$ autoregressive mode. Then the local residual energy
\begin{equation}
E_i(\hat{x}_i) := -\log q_i + \log p_i = -\log \frac{q_i}{p_i}
\label{eq:app_local_energy}
\end{equation}
is exactly the same discrepancy used in speculative decoding, since
\begin{equation}
\min\!\left(1,\frac{q_i}{p_i}\right) = \min(1, e^{-E_i(\hat{x}_i)}).
\label{eq:app_accept_energy}
\end{equation}
Thus, \ours{} and EDLM rely on the same AR-vs-diffusion discrepancy, but use it differently: EDLM uses it for global importance reweighting, while \ours{} uses it online through speculative acceptance and residual resampling.

\paragraph{NCE and a JSD-style interpretation.}
EDLM fits its AR-parameterized residual model using a noise-contrastive objective. Writing the EDLM NCE loss explicitly,
\begin{align}
\mathcal{L}_{\mathrm{NCE}}(\phi; \theta)
=&
-\mathbb{E}_{x_+ \sim q(\hat x_0 \mid x_t, x_0)}
\left[\log \sigma\!\left(\log \frac{p_{\phi}(x_+)}{p_{\theta}(x_+ \mid x_t)}\right)\right] \nonumber\\
&-
\mathbb{E}_{x_- \sim p_{\theta}(\hat x_0 \mid x_t)}
\left[\log \left(1-\sigma\!\left(\log \frac{p_{\phi}(x_-)}{p_{\theta}(x_- \mid x_t)}\right)\right)\right].
\label{eq:app_ddo}
\end{align}
This has the same binary likelihood-ratio form as Direct Discriminative Optimization (DDO) \citep{zheng2025direct}: positive samples are drawn from the true data posterior, while negative samples are drawn from the diffusion proposals. Under sufficient model capacity, the optimum is attained when the learned AR residual model matches the posterior. In this sense, the NCE objective admits a JSD-style interpretation through its binary likelihood-ratio form.

\paragraph{Relation to block-diffusion training.}
This perspective is complementary to standard diffusion training. Block-diffusion models are trained by optimizing the usual ELBO, which matches the reverse conditional to the posterior through a KL objective. In particular, the block-size-$1$ mode is not a separately trained energy model, but an extreme autoregressive conditional induced by the same diffusion model family. This does not imply that it exactly recovers the EDLM AR residual model. However, since both EDLM and block-diffusion training are ultimately driven by approximating the same posterior object from different variational viewpoints, the block-size-$1$ autoregressive mode provides a natural in-family proxy for an AR energy model.

\paragraph{A stochastic greedy interpretation.}
Finally, Eq.~\eqref{eq:app_accept_energy} suggests an optimization-style interpretation of speculative decoding. Since the acceptance probability is monotone in $-E_i(\hat{x}_i)$, drafted tokens with lower residual energy are more likely to be accepted, while higher-energy mismatches are more likely to be rejected and corrected by residual resampling. Therefore, although speculative decoding is not equivalent to exact energy minimization, it can be viewed heuristically as a stochastic, greedy local procedure that prefers lower-energy proposals. Under this view, \ours{} may be interpreted as an online local approximation to AR-guided energy correction.

\subsection{Evaluation setup details}
\label{app:eval_setup}

By default, we use \texttt{lm-eval-harness}~\citep{eval-harness}. We switch to our own scripts when task formatting or extraction robustness is critical.

\paragraph{GSM8K.}
For all models, we use our own script for more robust formatting handling and answer extraction than the default harness path. The system prompt is:
\begin{quote}
\small
\texttt{Solve the following math problem concisely and clearly and put your final answer within \textbackslash boxed\{\}.}
\end{quote}

\paragraph{MBPP and HumanEval.}
For \llada{} 2.1 on MBPP, we find the model does not reliably follow the \texttt{[BEGIN]} function-completion format expected by \texttt{lm-eval-harness}, which can lead to near-zero/zero measured accuracy if used directly. We therefore use our own script with more robust code extraction. The system prompt is:
\begin{quote}
\small
\texttt{You are an expert Python programmer. Write only the function code, no explanations.}
\end{quote}
For \sdar{}, we also evaluate HumanEval with our own script for the same formatting/extraction reason.

\paragraph{Summary of evaluation backends.}
Our own scripts are used for: GSM8K (all models), MBPP (\sdar{} and \llada{}2.1-Mini), and HumanEval (\sdar{}). Other settings use \texttt{lm-eval-harness}~\citep{eval-harness}. (By default, \texttt{lm-eval-harness} uses few-shot prompting on MBPP.)

\paragraph{Ablation subset.}
For ablations, we use the same fixed 200-sample subset of GSM8K across compared methods.

\subsection{Hyperparameter selection}
\label{app:hyperparameter_selection}

\ours{} introduces hyperparameters on two sides: model-side decoding parameters and routing-side verification parameters. On the model side, we keep confidence thresholds and remasking strategies at the default or recommended values of each model family whenever possible, and primarily vary the block structure: \sdar{} sweeps block sizes from $4$ to $64$, Fast-dLLM v2 fixes $B=32$ and sweeps sub-block sizes, and \llada{}2.1-Mini fixes $B=32$ while using the two default threshold settings reported in Section~\ref{sec:experiments}. On the routing side, we search over the policy family and its thresholds, including minimum-span thresholds, score thresholds, hysteresis on/off thresholds, penalty coefficients, and acceptance estimators.

For each model family, the search grid is the Cartesian product of the model-side and routing-side configurations. We evaluate configurations on two representative tasks, GSM8K and MBPP, using $100$ fixed examples from each task, and do not tune separately per downstream task. The full grid is run on the largest available model of each family (\sdar{}-8B, Fast-dLLM v2, and \llada{}2.1-Mini). For \sdar{}-1.7B and \sdar{}-4B, we use a reduced grid of approximately $14$ configurations selected from the \sdar{}-8B trends. The optional AR-like drafting/caching mask is also treated as a hyperparameter: it is enabled for \sdar{}, where it improves the tradeoff, but not for Fast-dLLM v2 or \llada{}2.1-Mini.

\subsection{Multi-seed \llada{}2.1-Mini results}
\label{app:llada_multiseed}

Table~\ref{tab:llada_multiseed} reports $5$-seed results on \llada{}2.1-Mini using seeds $\{0,1,2,3,42\}$ and $500$ examples. The deterministic greedy baselines are shown without standard deviations; \ours{} remains stochastic at temperature $0$ because of rejection sampling. These additional runs support the conclusion in Section~\ref{sec:experiments}: \ours{} is complementary to \llada{}2.1-Mini's built-in token editing, with the largest benefit appearing when sampling exposes the brittleness of aggressive diffusion decoding.

\begin{table*}[t]
\centering
\small
\renewcommand{\arraystretch}{1.12}
\setlength{\tabcolsep}{4pt}
\caption{Multi-seed \llada{}2.1-Mini results. Accuracy is reported in percent with wall-clock speedup over AR in parentheses.}
\label{tab:llada_multiseed}
\begin{tabular}{ll cccc}
\toprule
\textbf{Setting} & \textbf{Method} & \textbf{$T=0$ GSM8K} & \textbf{$T=0$ MBPP} & \textbf{$T=0.5$ GSM8K} & \textbf{$T=0.5$ MBPP} \\
\midrule
\multirow{3}{*}{Conservative}
& AR & 90.8 (1.0$\times$) & 65.8 (1.0$\times$) & $89.2{\pm}0.4$ (1.0$\times$) & $64.5{\pm}0.8$ (1.0$\times$) \\
& Diffusion & 91.0 (1.6$\times$) & 66.4 (1.7$\times$) & $89.8{\pm}0.7$ (2.4$\times$) & $56.5{\pm}0.3$ (2.9$\times$) \\
& \ours{} & $90.3{\pm}0.5$ (2.0$\times$) & $67.2{\pm}1.7$ (2.1$\times$) & $90.3{\pm}0.5$ (1.8$\times$) & $65.8{\pm}0.9$ (2.0$\times$) \\
\midrule
\multirow{3}{*}{Quality}
& AR & 90.8 (1.0$\times$) & 65.8 (1.0$\times$) & $89.2{\pm}0.4$ (1.0$\times$) & $64.5{\pm}0.8$ (1.0$\times$) \\
& Diffusion & 89.6 (2.5$\times$) & 57.8 (3.0$\times$) & $85.3{\pm}0.8$ (2.8$\times$) & $38.0{\pm}1.2$ (3.5$\times$) \\
& \ours{} & $89.8{\pm}0.6$ (1.9$\times$) & $64.8{\pm}1.0$ (2.2$\times$) & $89.0{\pm}0.7$ (1.8$\times$) & $60.8{\pm}1.1$ (2.1$\times$) \\
\bottomrule
\end{tabular}%
\end{table*}

\subsection{Throughput under batching}
\label{app:batch_throughput}

The main tables report wall-clock speedup at batch size $1$. We additionally measure throughput on $100$ GSM8K and $100$ MBPP examples at batch sizes $1$, $32$, and $64$, using a single NVIDIA H100 80GB. Throughput is measured both at a $128$-generated-token checkpoint and over the full generation. The same task examples are used for accuracy because routing decisions depend on the actual decoding trajectory.

\begin{table*}[t]
\centering
\small
\renewcommand{\arraystretch}{1.12}
\setlength{\tabcolsep}{4pt}
\caption{Batched throughput for \sdar{}-8B at $B=16,S=16$. Throughput is reported in tokens/s.}
\label{tab:batch_throughput_sdar}
\begin{tabular}{c l ccc ccc}
\toprule
\multirow{2}{*}{\textbf{BS}} & \multirow{2}{*}{\textbf{Method}} & \multicolumn{3}{c}{\textbf{GSM8K}} & \multicolumn{3}{c}{\textbf{MBPP}} \\
\cmidrule(lr){3-5}\cmidrule(lr){6-8}
& & Acc & Tput@128 & Tput full & Acc & Tput@128 & Tput full \\
\midrule
1  & \bdthree{} dynamic & 0.87 & 38 & 125 & 0.43 & 33 & 275 \\
1  & \ours{} & 0.86 & 45 & 143 & 0.72 & 46 & 354 \\
32 & \bdthree{} dynamic & 0.85 & 1064 & 3458 & 0.41 & 902 & 7485 \\
32 & \ours{} & 0.85 & 1378 & 4555 & 0.69 & 1195 & 8914 \\
64 & \bdthree{} dynamic & 0.85 & 1800 & 5809 & 0.40 & 1491 & 12092 \\
64 & \ours{} & 0.86 & 1825 & 5962 & 0.70 & 1601 & 11950 \\
\bottomrule
\end{tabular}%
\end{table*}

\begin{table*}[t]
\centering
\small
\renewcommand{\arraystretch}{1.12}
\setlength{\tabcolsep}{4pt}
\caption{Batched throughput for Fast-dLLM v2 at $B=32,\mathrm{SB}=32$. Throughput is reported in tokens/s.}
\label{tab:batch_throughput_fast}
\begin{tabular}{c l ccc ccc}
\toprule
\multirow{2}{*}{\textbf{BS}} & \multirow{2}{*}{\textbf{Method}} & \multicolumn{3}{c}{\textbf{GSM8K}} & \multicolumn{3}{c}{\textbf{MBPP}} \\
\cmidrule(lr){3-5}\cmidrule(lr){6-8}
& & Acc & Tput@128 & Tput full & Acc & Tput@128 & Tput full \\
\midrule
1  & Baseline & 0.78 & 146 & 342 & 0.41 & 83 & 237 \\
1  & \ours{} & 0.89 & 143 & 310 & 0.47 & 113 & 327 \\
32 & Baseline & 0.83 & 2021 & 4515 & 0.44 & 1120 & 3215 \\
32 & \ours{} & 0.89 & 2535 & 5160 & 0.51 & 1715 & 4757 \\
64 & Baseline & 0.83 & 2137 & 4752 & 0.44 & 1181 & 3385 \\
64 & \ours{} & 0.89 & 2653 & 5386 & 0.51 & 1822 & 5038 \\
\bottomrule
\end{tabular}%
\end{table*}

The batched results are consistent with the single-example trends. On Fast-dLLM v2, the NFE reduction translates directly into throughput gains because verification does not require the 2$L$ construction. On \sdar{}, high-batch throughput gains are more modest because the 2$L$ verifier increases the effective verifier workload, but the MBPP accuracy improvement remains large across all batch sizes.

\subsection{Failure cases and boundary conditions}
\label{app:failure_boundary}

\ours{} is most useful when confidence thresholding alone misses local sequence-level inconsistencies while the block-size-$1$ AR mode remains a reliable verifier. Table~\ref{tab:confidence_acceptance_quadrants} categorizes verified tokens by draft confidence $p_{\mathrm{sel}}$ and verifier-to-draft ratio $q/p$, using $0.5$ as the threshold for both axes. The overconfident quadrant, high draft confidence but low verifier ratio, is small but important: these are tokens that pure confidence thresholding would tend to commit but \ours{} usually rejects. Conversely, the low-confidence/high-ratio quadrant shows where verification can accept useful tokens that confidence thresholding would otherwise remask.

\begin{table}[t]
\centering
\small
\renewcommand{\arraystretch}{1.12}
\setlength{\tabcolsep}{5pt}
\caption{Per-token draft-confidence versus verifier-ratio analysis on GSM8K. Fractions are over verified tokens.}
\label{tab:confidence_acceptance_quadrants}
\begin{tabular}{lccc}
\toprule
\textbf{Quadrant} & \textbf{Description} & \textbf{\sdar{}} & \textbf{Fast-dLLM v2} \\
\midrule
Q1 & High conf., high ratio & 47.9\%, 99.2\% acc. & 44.6\%, 99.3\% acc. \\
Q2 & Low conf., high ratio & 28.5\%, 97.8\% acc. & 21.3\%, 96.4\% acc. \\
Q3 & High conf., low ratio & 2.4\%, 5.7\% acc. & 3.8\%, 4.6\% acc. \\
Q4 & Low conf., low ratio & 21.2\%, 5.4\% acc. & 30.3\%, 4.8\% acc. \\
\bottomrule
\end{tabular}
\end{table}

A clear boundary condition appears when the block-size-$1$ AR mode is weak. Table~\ref{tab:d2f_failure} evaluates a D2F \citep{wang2025diffusion}  \llada{} variant on $30$ GSM8K examples. Its AR mode substantially underperforms diffusion decoding, so residual resampling from the AR verifier can reduce accuracy. Disabling residual resampling and allowing diffusion to re-denoise rejected positions largely recovers the diffusion result. Thus, \ours{} should be used cautiously when the model is not trained to preserve a reliable blockwise AR conditional.

\begin{table}[t]
\centering
\small
\renewcommand{\arraystretch}{1.12}
\caption{Failure case on a D2F \llada{} variant evaluated on GSM8K.}
\label{tab:d2f_failure}
\begin{tabular}{lcc}
\toprule
\textbf{Config} & \textbf{Accuracy} & \textbf{Avg. NFE} \\
\midrule
AR ($B=1$) & 20.0\% & 225.5 \\
Diffusion ($B=32$) & 50.0\% & 72.6 \\
\ours{} ($B=32$, resampling) & 33.3\% & 58.5 \\
\ours{} ($B=32$, reject only) & 46.7\% & 67.7 \\
\bottomrule
\end{tabular}
\end{table}

\subsection{Inference-time alternatives}
\label{app:inference_alternatives}

We also compare with two classes of inference-time alternatives: multi-sample importance reweighting and single-sample correction schedules. For the first class, we implement self-normalized importance sampling with $k=8$ candidates, using either EDLM-style residual energy or NLL as the energy. Table~\ref{tab:snis_llada} shows that importance reweighting can improve some \llada{}2.1-Mini settings, but it requires substantially more compute because candidate generation and reranking add extra passes. On \sdar{}-8B, the same procedure did not provide consistent gains in our runs.

\begin{table*}[t]
\centering
\small
\renewcommand{\arraystretch}{1.12}
\setlength{\tabcolsep}{5pt}
\caption{Self-normalized importance sampling on \llada{}2.1-Mini with $k=8$. Accuracy is GSM8K/MBPP. Slowdown is approximate relative to single-sample baseline generation.}
\label{tab:snis_llada}
\begin{tabular}{lcccc}
\toprule
\textbf{Config} & \textbf{$B=8$} & \textbf{$B=16$} & \textbf{$B=32$} & \textbf{Slowdown} \\
\midrule
Baseline & 85.0 / 61.5 & 90.5 / 63.5 & 90.0 / 56.0 & 1$\times$ \\
Residual, $w=0.2S$ & 84.0 / 60.5 & 88.5 / 61.0 & 92.5 / 61.0 & $\sim$3$\times$ \\
Residual, $w=S$ & 84.5 / 63.5 & 91.0 / 59.0 & 88.5 / 57.5 & $\sim$10$\times$ \\
NLL, $w=0.2S$ & 85.5 / 63.0 & 90.5 / 58.5 & 91.0 / 57.5 & $\sim$3$\times$ \\
NLL, $w=S$ & 90.5 / 63.0 & 92.0 / 61.5 & 92.5 / 59.0 & $\sim$10$\times$ \\
\bottomrule
\end{tabular}%
\end{table*}

For the second class, Table~\ref{tab:alternative_schedules} compares \ours{} against periodic AR correction and stronger confidence schedules on \sdar{}-8B with $B=16$. These baselines improve over some aggressive confidence-threshold settings, but they either lose more accuracy at similar speed or require more NFE to approach \ours{}. In this setting, \ours{} gives the strongest accuracy--efficiency tradeoff among the tested single-sample alternatives.

\begin{table*}[t]
\centering
\small
\renewcommand{\arraystretch}{1.12}
\setlength{\tabcolsep}{5pt}
\caption{Inference-time alternatives on \sdar{}-8B with $B=16$. Speedups are relative to AR.}
\label{tab:alternative_schedules}
\begin{tabular}{lccc ccc}
\toprule
\multirow{2}{*}{\textbf{Method}} & \multicolumn{3}{c}{\textbf{GSM8K}} & \multicolumn{3}{c}{\textbf{MBPP}} \\
\cmidrule(lr){2-4}\cmidrule(lr){5-7}
& Acc & NFE & Wall & P@1 & NFE & Wall \\
\midrule
AR ($B=1$) & 0.900 & 1.00$\times$ & 1.00$\times$ & 0.665 & 1.00$\times$ & 1.00$\times$ \\
\ours{} config-B & 0.895 & 4.65$\times$ & 3.62$\times$ & 0.595 & 3.77$\times$ & 3.04$\times$ \\
\bdthree{} dynamic $\tau=0.85$ & 0.835 & 3.63$\times$ & 3.58$\times$ & 0.465 & 2.92$\times$ & 2.73$\times$ \\
Periodic step $p=2$ & 0.865 & 4.33$\times$ & 3.28$\times$ & 0.480 & 3.29$\times$ & 2.58$\times$ \\
Periodic block $bp=2$ & 0.835 & 4.47$\times$ & 3.71$\times$ & 0.490 & 3.43$\times$ & 2.74$\times$ \\
Conf. H$\to$L $0.95{\to}0.5$ & 0.860 & 3.47$\times$ & 3.36$\times$ & 0.475 & 2.64$\times$ & 2.61$\times$ \\
Conf. L$\to$H $0.5{\to}0.95$ & 0.710 & 4.93$\times$ & 4.35$\times$ & 0.250 & 3.98$\times$ & 3.87$\times$ \\
Margin $\tau=0.7$ & 0.825 & 3.91$\times$ & 3.31$\times$ & 0.400 & 3.09$\times$ & 2.73$\times$ \\
\bottomrule
\end{tabular}%
\end{table*}

\subsection{Additional ablations on token acceptance estimators}
\label{app:estimator_ablation}

We further ablate several lightweight estimators for the accepted prefix length predictor in Eq.~\eqref{eq:khat}. In addition to the two estimators used in the main text, we test four additional variants and a random baseline. Results are shown in Table~\ref{tab:token_acceptance_estimator_ablation}.

\paragraph{Setup.}
We evaluate these estimators on \sdar{}-8B-Chat using 50 samples from GSM8K and 50 from MBPP. We invoke speculative decoding at every denoising step and compare the estimated accepted prefix length $\hat K$ against the actual accepted prefix length under verification. We report signed error, error standard deviation, and mean absolute error (MAE).

\paragraph{Findings.}
Hard margin thresholding gives the most accurate estimates. In particular, $\alpha_i = \mathbf{1}[m_i \ge \tau_{\mathrm{margin}}]$ with $\tau_{\mathrm{margin}}=0.1$ achieves the lowest MAE in Table~\ref{tab:token_acceptance_estimator_ablation}. However, margin-based estimators are more sensitive to calibration, and the best threshold can vary across models and sampling settings. Moreover, better estimation accuracy does not necessarily imply better downstream decoding accuracy. Although hard margin thresholding is the most accurate estimator in Table~\ref{tab:token_acceptance_estimator_ablation}, the soft entropy-based estimator yields higher average task accuracy across block sizes in Tables~\ref{tab:ablation_policy_sdar_score_entropy} and~\ref{tab:ablation_policy_sdar_score_margin}.
For this reason, we use the soft entropy-based estimator, $\alpha_i = \exp(-\beta \tilde H_i)$, in the main experiments. It is simpler to use, less tied to model-specific calibration, and empirically more effective in our routing setup. Table~\ref{tab:token_acceptance_estimator_ablation} also suggests that more accurate, better-calibrated estimators may yield further gains.

\begin{table}[t]
\centering
\small
\renewcommand{\arraystretch}{1.12}
\setlength{\tabcolsep}{6pt}
\caption{Token acceptance estimator ablation. We report signed error (Mean Error), error standard deviation (Std Error), and mean absolute error (MAE) for predicting accepted prefix length. Lower absolute error is better.}
\label{tab:token_acceptance_estimator_ablation}
\begin{tabular}{lccc}
\toprule
\textbf{Config} & \textbf{Mean Error} & \textbf{Std Error} & \textbf{MAE} \\
\midrule

\multicolumn{4}{c}{\textbf{Random baseline} \(\alpha_i \sim \mathcal{U}[0,1]\)} \\
\midrule
-- & -8.014 & 10.293 & 8.129 \\
\midrule

\multicolumn{4}{c}{\textbf{Soft entropy estimator} \(\alpha_i = \exp(-\beta \tilde H_i)\)} \\
\midrule
\(\beta=0.5\)  & -0.445 & 8.598 & 5.843 \\
\(\beta=0.75\) & -1.956 & 8.872 & 5.279 \\
\(\beta=1.0\)  & -2.861 & 9.031 & 5.084 \\
\(\beta=1.25\) & -3.471 & 9.129 & 5.059 \\
\(\beta=1.5\)  & -3.686 & 8.969 & 4.901 \\
\midrule

\multicolumn{4}{c}{\textbf{Confidence-power estimator} \(\alpha_i = p_i^{\gamma_{\mathrm{conf}}}\)} \\
\midrule
\(\gamma_{\mathrm{conf}}=0.5\)  & -4.107 & 8.220 & 4.660 \\
\(\gamma_{\mathrm{conf}}=0.75\) & -4.838 & 8.543 & 5.113 \\
\(\gamma_{\mathrm{conf}}=1.0\)  & -5.372 & 8.822 & 5.564 \\
\(\gamma_{\mathrm{conf}}=1.25\) & -5.528 & 8.787 & 5.654 \\
\(\gamma_{\mathrm{conf}}=1.5\)  & -5.724 & 8.824 & 5.825 \\
\midrule

\multicolumn{4}{c}{\textbf{Rényi-2 entropy estimator} \(\alpha_i = \sum_v p_i(v)^2\)} \\
\midrule
-- & -5.870 & 8.922 & 5.961 \\
\midrule

\multicolumn{4}{c}{\textbf{Hard-entropy threshold} \(\alpha_i = \mathbf{1}[\tilde H_i < \tau_{\mathrm{ent}}]\)} \\
\midrule
\(\tau_{\mathrm{ent}}=0.01\) & -7.631 & 9.489 & 7.634 \\
\(\tau_{\mathrm{ent}}=0.02\) & -7.271 & 9.425 & 7.276 \\
\(\tau_{\mathrm{ent}}=0.05\) & -6.505 & 9.305 & 6.535 \\
\(\tau_{\mathrm{ent}}=0.1\)  & -5.516 & 9.260 & 5.732 \\
\(\tau_{\mathrm{ent}}=0.2\)  & -3.333 & 8.931 & 4.835 \\
\midrule

\multicolumn{4}{c}{\textbf{Hard-margin threshold} \(\alpha_i = \mathbf{1}[m_i \ge \tau_{\mathrm{margin}}]\)} \\
\midrule
\(\tau_{\mathrm{margin}}=0.01\) & 4.240 & 5.871 & 4.367 \\
\(\tau_{\mathrm{margin}}=0.02\) & 2.593 & 4.359 & 2.868 \\
\(\tau_{\mathrm{margin}}=0.05\) & 0.953 & 2.898 & 1.599 \\
\(\tau_{\mathrm{margin}}=0.1\)  & 0.065 & 2.476 & \textbf{1.245} \\
\(\tau_{\mathrm{margin}}=0.2\)  & -0.823 & 2.829 & 1.371 \\
\bottomrule
\end{tabular}
\end{table}

\subsection{Contextual-bandit routing policy}
\label{app:bandit_policy}

As an additional adaptive routing baseline, we implement a simple contextual-bandit policy based on upper confidence bounds (UCB). At each decoding decision, the policy chooses between two actions, $a \in \{0,1\}$, where $a=0$ denotes standard diffusion decoding without verification and $a=1$ denotes invoking speculative verification.

Let $b_t$ denote the context bucket at decision step $t$. The UCB policy selects
\begin{equation}
a_t = \arg\max_{a}\left(\hat{\mu}_{a,b_t} + \beta \sqrt{\frac{\log t}{n_{a,b_t}}}\right),
\label{eq:ucb}
\end{equation}
where $\hat{\mu}_{a,b}$ is the empirical mean reward of action $a$ in context bucket $b$, $n_{a,b}$ is the number of times that action has been taken in that bucket, and $\beta$ controls exploration. The empirical mean reward is computed as
\begin{equation}
\hat{\mu}_{a,b} = \frac{1}{n_{a,b}} \sum_{s:\, a_s=a,\; b_s=b} r_s, \qquad
n_{a,b} = \#\{s : a_s=a,\; b_s=b\}.
\end{equation}
Following our implementation, the reward is defined as
\begin{equation}
r = \frac{\texttt{decoded\_this\_step}}{\texttt{time\_cost}}, \qquad
\texttt{time\_cost} =
\begin{cases}
2, & \text{if verification is invoked},\\
1, & \text{otherwise}.
\end{cases}
\end{equation}

We use a simple discretized context tuple consisting of: (i) span-length bin for the first contiguous masked span, linearly partitioned from $1$ to block size; (ii) decoding-progress bin, measured by the fraction of already unmasked tokens in the current block; and (iii) entropy bin, computed from normalized token entropy. By default, we use two bins for each dimension. We report the ablation results for different context tuples and exploration coefficients in Table~\ref{tab:ablation_policy_sdar_ucb}.
In our experiments, this contextual-bandit policy serves as a representative simple training-free reinforcement learning baseline. Although adaptive, it is more involved than the threshold-based policies and was not the best-performing routing strategy in our setting.

\subsection{Ablation on routing policies}
\label{app:routing_ablation}

We provide a comprehensive ablation of routing policies on \sdar{}-8B-Chat over GSM8K and MBPP, using 200 samples from each dataset. We evaluate block sizes $B \in \{4,8,16,32\}$. The corresponding \bdthree{} baseline \citep{arriola2025block} results are reported in Table~\ref{tab:baseline_bd3}.

\paragraph{Minimum-span policy.}
Table~\ref{tab:ablation_policy_sdar_span} reports the ablation of the minimum-span policy, where verification is invoked whenever the first contiguous masked span satisfies $|C_t| \ge \tau_{\mathrm{span}}$. We sweep $\tau_{\mathrm{span}} \in \{1,2,4,\dots,B-1\}$. Figure~\ref{fig:sdar8b_acc_speed} further plots the cases $\tau_{\mathrm{span}} \in \{1,2\}$ together with different denoising steps $S$, showing that \ours{} generally lies in the upper-left region of the accuracy--speed tradeoff.

\paragraph{Score-threshold policy.}
For the score-threshold policy, we study two choices of accepted-prefix estimator. Table~\ref{tab:ablation_policy_sdar_score_entropy} uses the soft entropy-based estimator ($\beta=1$), while Table~\ref{tab:ablation_policy_sdar_score_margin} uses the hard confidence-margin estimator ($\tau_\text{margin}=0.05$). Across most settings, both score-based variants outperform the \bdthree{} baseline in accuracy under the same block size, with the gains being especially pronounced at larger block sizes $\mathrm{B}=16$ and $\mathrm{B}=32$.

Comparing the two estimators, the entropy-based estimator consistently achieves higher downstream accuracy than the margin-based estimator across all block sizes. This is notable because, as shown in Table~\ref{tab:token_acceptance_estimator_ablation}, the hard margin estimator is more accurate at predicting accepted prefix length.

\paragraph{Hysteresis policy.}
Table~\ref{tab:ablation_policy_sdar_hysteresis} reports the ablation of the hysteresis-based routing strategy. Compared with the plain score-threshold policy, hysteresis introduces temporal consistency by using different thresholds for turning verification on and off.

\paragraph{Contextual-bandit policy.}
Finally, Table~\ref{tab:ablation_policy_sdar_ucb} reports the UCB-style contextual-bandit routing policy. More implementation details are given in Appendix~\ref{app:bandit_policy}.

\begin{table*}[ht]
\centering
\small
\renewcommand{\arraystretch}{1.12}
\caption{Baseline BD3 results with static and dynamic confidence-based unmasking.}
\label{tab:baseline_bd3}
\resizebox{\linewidth}{!}{%
\begin{tabular}{c ccc ccc ccc ccc ccc}
\toprule
\multirow{2}{*}{Config} & \multicolumn{3}{c}{\textbf{B=1} (AR)} & \multicolumn{3}{c}{\textbf{B=4}} & \multicolumn{3}{c}{\textbf{B=8}} & \multicolumn{3}{c}{\textbf{B=16}} & \multicolumn{3}{c}{\textbf{B=32}} \\
\cmidrule(lr){2-4}\cmidrule(lr){5-7}\cmidrule(lr){8-10}\cmidrule(lr){11-13}\cmidrule(lr){14-16}
& GSM8K & MBPP & Avg & GSM8K & MBPP & Avg & GSM8K & MBPP & Avg & GSM8K & MBPP & Avg & GSM8K & MBPP & Avg \\
\midrule
Static
& 89.0 (1.0$\times$) & 63.6 (1.0$\times$) & 76.3 (1.0$\times$)
& 90.5 (1.5$\times$) & \textbf{62.6} (1.5$\times$) & \textbf{76.6} (1.5$\times$)
& 90.5 (1.8$\times$) & 50.8 (1.7$\times$) & 70.7 (1.7$\times$)
& \textbf{89.0} (1.6$\times$) & 36.2 (1.7$\times$) & 62.6 (1.6$\times$)
& \textbf{86.0} (1.4$\times$) & 12.4 (1.5$\times$) & 49.2 (1.4$\times$) \\
Dynamic
& -- & -- & --
& \textbf{91.0} (\textbf{2.9}$\times$) & 60.8 (\textbf{2.0}$\times$) & 75.9 (\textbf{2.4}$\times$)
& 90.5 (\textbf{3.3}$\times$) & \textbf{53.6} (\textbf{2.6}$\times$) & \textbf{72.0} (\textbf{3.0}$\times$)
& 86.0 (\textbf{3.4}$\times$) & \textbf{49.0} (\textbf{2.8}$\times$) & \textbf{67.5} (\textbf{3.1}$\times$)
& 83.0 (\textbf{3.2}$\times$) & \textbf{43.6} (\textbf{2.8}$\times$) & \textbf{63.3} (\textbf{3.0}$\times$) \\
\bottomrule
\end{tabular}%
}
\end{table*}

\begin{table*}[ht]
\centering
\small
\renewcommand{\arraystretch}{1.12}
\caption{Ablation of the minimum-span policy. We vary the minimum contiguous mask-span threshold \(\tau_{\mathrm{span}}\).}
\label{tab:ablation_policy_sdar_span}
\resizebox{\linewidth}{!}{%
\begin{tabular}{c ccc ccc ccc ccc}
\toprule
\(\tau_{\mathrm{span}}\) & \multicolumn{3}{c}{\textbf{B=4}} & \multicolumn{3}{c}{\textbf{B=8}} & \multicolumn{3}{c}{\textbf{B=16}} & \multicolumn{3}{c}{\textbf{B=32}} \\
\cmidrule(lr){2-4}\cmidrule(lr){5-7}\cmidrule(lr){8-10}\cmidrule(lr){11-13}
& GSM8K & MBPP & Avg & GSM8K & MBPP & Avg & GSM8K & MBPP & Avg & GSM8K & MBPP & Avg \\
\midrule
1  & 90.0 (1.9$\times$) & \textbf{61.8} (1.9$\times$) & 75.9 (1.9$\times$) & 90.0 (2.9$\times$) & \textbf{61.2} (2.5$\times$) & 75.6 (2.7$\times$) & 90.0 (3.5$\times$) & \textbf{61.4} (\textbf{3.0}$\times$) & \textbf{75.7} (3.2$\times$) & 89.5 (4.7$\times$) & 59.8 (3.2$\times$) & 74.6 (3.8$\times$) \\
2  & 91.5 (\textbf{2.1}$\times$) & 59.6 (\textbf{2.0}$\times$) & 75.6 (\textbf{2.0}$\times$) & \textbf{92.0} (3.2$\times$) & 60.6 (2.5$\times$) & \textbf{76.3} (2.8$\times$) & 90.0 (3.9$\times$) & 61.0 (2.9$\times$) & 75.5 (3.3$\times$) & \textbf{93.0} (\textbf{4.8}$\times$) & 58.6 (\textbf{3.4}$\times$) & 75.8 (\textbf{4.0}$\times$) \\
3  & \textbf{93.0} (1.9$\times$) & 61.2 (1.9$\times$) & \textbf{77.1} (1.9$\times$) & -- & -- & -- & -- & -- & -- & -- & -- & -- \\
4  & -- & -- & -- & 91.5 (\textbf{3.3}$\times$) & 60.4 (\textbf{2.8}$\times$) & 76.0 (\textbf{3.0}$\times$) & \textbf{91.5} (4.1$\times$) & 59.4 (\textbf{3.0}$\times$) & 75.4 (\textbf{3.5}$\times$) & 91.5 (\textbf{4.8}$\times$) & 58.8 (3.1$\times$) & 75.2 (3.8$\times$) \\
7  & -- & -- & -- & 91.5 (3.2$\times$) & 58.8 (2.6$\times$) & 75.2 (2.9$\times$) & -- & -- & -- & -- & -- & -- \\
8  & -- & -- & -- & -- & -- & -- & 90.0 (\textbf{4.3}$\times$) & 58.0 (2.8$\times$) & 74.0 (3.4$\times$) & 92.0 (4.7$\times$) & \textbf{61.0} (3.2$\times$) & \textbf{76.5} (3.8$\times$) \\
15 & -- & -- & -- & -- & -- & -- & 88.5 (4.1$\times$) & 56.8 (2.7$\times$) & 72.6 (3.3$\times$) & -- & -- & -- \\
16 & -- & -- & -- & -- & -- & -- & -- & -- & -- & 85.5 (4.1$\times$) & 55.2 (2.9$\times$) & 70.3 (3.4$\times$) \\
31 & -- & -- & -- & -- & -- & -- & -- & -- & -- & 84.5 (3.8$\times$) & 50.4 (2.8$\times$) & 67.5 (3.2$\times$) \\
\bottomrule
\end{tabular}%
}
\end{table*}

\begin{table*}[ht]
\centering
\small
\renewcommand{\arraystretch}{1.12}
\caption{Ablation of score-threshold policies with \textbf{entropy}-based token acceptance estimator. Static score uses \(s=\hat K-c\); dynamic score uses \(s=\hat K-c\cdot N_{\mathrm{hi}}\).}
\label{tab:ablation_policy_sdar_score_entropy}
\resizebox{\linewidth}{!}{%
\begin{tabular}{cc ccc ccc ccc ccc}
\toprule
\(\tau_{\mathrm{score}}\) & \(c\) & \multicolumn{3}{c}{\textbf{B=4}} & \multicolumn{3}{c}{\textbf{B=8}} & \multicolumn{3}{c}{\textbf{B=16}} & \multicolumn{3}{c}{\textbf{B=32}} \\
\cmidrule(lr){3-5}\cmidrule(lr){6-8}\cmidrule(lr){9-11}\cmidrule(lr){12-14}
& & GSM8K & MBPP & Avg & GSM8K & MBPP & Avg & GSM8K & MBPP & Avg & GSM8K & MBPP & Avg \\
\midrule
\multicolumn{14}{c}{\textbf{Static score} \(s=\hat K-c\)} \\
\midrule
\multirow{3}{*}{-5}
& 1 & 90.0 (2.0$\times$) & 62.0 (1.9$\times$) & 76.0 (2.0$\times$) & 92.5 (3.3$\times$) & \textbf{62.6} (2.6$\times$) & \textbf{77.6} (2.9$\times$) & 90.0 (3.6$\times$) & 59.0 (2.9$\times$) & 74.5 (3.2$\times$) & 89.5 (\textbf{4.7}$\times$) & 59.8 (\textbf{3.4}$\times$) & 74.6 (\textbf{4.0}$\times$) \\
& 2 & 93.0 (2.1$\times$) & 62.0 (2.0$\times$) & 77.5 (2.1$\times$) & 91.5 (2.8$\times$) & \textbf{62.6} (2.4$\times$) & 77.0 (2.6$\times$) & 91.0 (4.0$\times$) & 59.8 (3.2$\times$) & 75.4 (3.6$\times$) & 90.0 (4.2$\times$) & 57.0 (3.2$\times$) & 73.5 (3.7$\times$) \\
& 4 & 93.0 (2.3$\times$) & 61.8 (2.0$\times$) & 77.4 (2.1$\times$) & 91.5 (3.0$\times$) & 61.2 (2.5$\times$) & 76.4 (2.8$\times$) & 90.5 (\textbf{4.2}$\times$) & 61.4 (3.0$\times$) & 76.0 (3.5$\times$) & 90.0 (3.9$\times$) & 59.8 (3.0$\times$) & 74.9 (3.4$\times$) \\
\cmidrule(lr){1-14}
\multirow{3}{*}{-1}
& 1 & 90.0 (2.2$\times$) & 62.0 (2.0$\times$) & 76.0 (2.1$\times$) & 90.0 (3.0$\times$) & 61.2 (2.5$\times$) & 75.6 (2.7$\times$) & 90.5 (\textbf{4.2}$\times$) & 61.4 (3.2$\times$) & 76.0 (\textbf{3.7}$\times$) & 90.0 (4.4$\times$) & 59.8 (3.1$\times$) & 74.9 (3.7$\times$) \\
& 2 & 92.0 (2.0$\times$) & 62.2 (1.9$\times$) & 77.1 (2.0$\times$) & 90.5 (3.3$\times$) & 61.8 (2.8$\times$) & 76.2 (3.0$\times$) & 89.5 (3.8$\times$) & 59.4 (3.0$\times$) & 74.4 (3.3$\times$) & 92.0 (4.7$\times$) & \textbf{60.8} (3.4$\times$) & 76.4 (4.0$\times$) \\
& 4 & 89.5 (2.0$\times$) & \textbf{63.8} (2.0$\times$) & 76.6 (2.0$\times$) & \textbf{94.0} (3.4$\times$) & 60.6 (2.8$\times$) & 77.3 (3.1$\times$) & 91.0 (3.6$\times$) & \textbf{62.4} (2.8$\times$) & \textbf{76.7} (3.2$\times$) & 87.5 (4.5$\times$) & 58.2 (3.3$\times$) & 72.8 (3.9$\times$) \\
\cmidrule(lr){1-14}
\multirow{3}{*}{0}
& 1 & 92.0 (2.3$\times$) & 62.2 (2.0$\times$) & 77.1 (2.1$\times$) & 88.5 (3.3$\times$) & 61.8 (2.7$\times$) & 75.2 (3.0$\times$) & 90.5 (3.6$\times$) & 59.4 (2.9$\times$) & 75.0 (3.2$\times$) & \textbf{93.0} (4.5$\times$) & \textbf{60.8} (3.4$\times$) & \textbf{76.9} (3.9$\times$) \\
& 2 & \textbf{93.5} (2.3$\times$) & 62.4 (2.1$\times$) & \textbf{78.0} (2.2$\times$) & 92.0 (3.2$\times$) & 58.6 (2.4$\times$) & 75.3 (2.8$\times$) & \textbf{93.0} (\textbf{4.2}$\times$) & 58.2 (3.1$\times$) & 75.6 (3.6$\times$) & 90.5 (4.3$\times$) & 59.4 (3.1$\times$) & 75.0 (3.6$\times$) \\
& 4 & 90.0 (2.5$\times$) & 59.4 (2.2$\times$) & 74.7 (2.3$\times$) & 91.5 (3.3$\times$) & 59.8 (2.5$\times$) & 75.6 (2.9$\times$) & 90.5 (4.1$\times$) & 58.8 (3.2$\times$) & 74.6 (3.6$\times$) & 91.5 (4.2$\times$) & 56.2 (3.0$\times$) & 73.9 (3.5$\times$) \\
\cmidrule(lr){1-14}
\multirow{3}{*}{1}
& 1 & 90.5 (2.3$\times$) & 62.4 (2.0$\times$) & 76.4 (2.2$\times$) & 92.0 (3.2$\times$) & 57.2 (2.7$\times$) & 74.6 (2.9$\times$) & 91.0 (3.9$\times$) & 58.2 (3.2$\times$) & 74.6 (3.5$\times$) & 90.5 (4.3$\times$) & 59.4 (3.1$\times$) & 75.0 (3.6$\times$) \\
& 2 & 89.5 (2.1$\times$) & 61.2 (1.9$\times$) & 75.4 (2.0$\times$) & 91.0 (3.5$\times$) & 61.6 (2.8$\times$) & 76.3 (3.1$\times$) & 91.0 (3.8$\times$) & \textbf{62.4} (3.0$\times$) & \textbf{76.7} (3.4$\times$) & 89.5 (4.7$\times$) & 60.2 (3.3$\times$) & 74.8 (3.9$\times$) \\
& 4 & 90.5 (2.2$\times$) & 59.4 (2.0$\times$) & 75.0 (2.1$\times$) & 91.0 (3.5$\times$) & 56.8 (2.9$\times$) & 73.9 (3.2$\times$) & 90.0 (3.9$\times$) & 53.4 (2.9$\times$) & 71.7 (3.3$\times$) & 87.5 (3.6$\times$) & 55.2 (3.2$\times$) & 71.4 (3.4$\times$) \\
\cmidrule(lr){1-14}
\multirow{3}{*}{5}
& 1 & 90.5 (2.8$\times$) & 60.6 (1.9$\times$) & 75.6 (2.2$\times$) & 93.0 (3.3$\times$) & 56.2 (2.9$\times$) & 74.6 (3.1$\times$) & 90.5 (3.7$\times$) & 53.8 (2.9$\times$) & 72.2 (3.2$\times$) & 89.5 (3.5$\times$) & 53.0 (3.0$\times$) & 71.2 (3.2$\times$) \\
& 2 & 91.0 (\textbf{2.9}$\times$) & 60.6 (2.2$\times$) & 75.8 (2.5$\times$) & 91.5 (2.9$\times$) & 55.0 (2.6$\times$) & 73.2 (2.8$\times$) & 86.5 (3.7$\times$) & 51.6 (2.9$\times$) & 69.0 (3.3$\times$) & 88.0 (3.2$\times$) & 48.6 (2.6$\times$) & 68.3 (2.9$\times$) \\
& 4 & 91.0 (\textbf{2.9}$\times$) & 60.6 (2.2$\times$) & 75.8 (2.5$\times$) & 90.0 (3.2$\times$) & 53.4 (2.7$\times$) & 71.7 (2.9$\times$) & 88.0 (3.4$\times$) & 49.4 (2.8$\times$) & 68.7 (3.1$\times$) & 83.5 (2.9$\times$) & 44.8 (2.6$\times$) & 64.1 (2.7$\times$) \\
\midrule
\multicolumn{14}{c}{\textbf{Dynamic score} \(s=\hat K-c\cdot N_{\mathrm{hi}}\)} \\
\midrule
-5 & 1 & 92.0 (2.0$\times$) & 62.0 (1.8$\times$) & 77.0 (1.9$\times$) & \textbf{92.5} (3.0$\times$) & \textbf{62.6} (2.5$\times$) & \textbf{77.6} (2.8$\times$) & 90.0 (3.8$\times$) & 59.0 (2.9$\times$) & 74.5 (3.3$\times$) & 91.0 (4.5$\times$) & 58.2 (3.4$\times$) & 74.6 (3.9$\times$) \\
-1 & 1 & 89.0 (2.2$\times$) & 62.0 (2.0$\times$) & 75.5 (2.1$\times$) & 91.5 (2.8$\times$) & 61.2 (2.4$\times$) & 76.4 (2.6$\times$) & 90.0 (3.9$\times$) & \textbf{60.0} (3.2$\times$) & 75.0 (3.5$\times$) & \textbf{92.0} (4.1$\times$) & 58.4 (3.0$\times$) & \textbf{75.2} (3.5$\times$) \\
0  & 1 & \textbf{93.5} (2.4$\times$) & 60.6 (2.1$\times$) & 77.0 (2.3$\times$) & 90.5 (\textbf{3.8}$\times$) & 59.6 (\textbf{2.8}$\times$) & 75.0 (\textbf{3.3}$\times$) & 90.5 (\textbf{4.1}$\times$) & \textbf{60.0} (2.9$\times$) & \textbf{75.2} (3.4$\times$) & 89.0 (\textbf{4.8}$\times$) & \textbf{59.4} (\textbf{3.5}$\times$) & 74.2 (\textbf{4.1}$\times$) \\
1  & 1 & 92.5 (\textbf{2.9}$\times$) & \textbf{62.6} (\textbf{2.4}$\times$) & \textbf{77.6} (\textbf{2.6}$\times$) & 86.0 (3.4$\times$) & 57.6 (2.6$\times$) & 71.8 (3.0$\times$) & \textbf{92.0} (3.7$\times$) & 58.6 (\textbf{3.3}$\times$) & 74.3 (\textbf{3.7}$\times$) & 91.5 (4.3$\times$) & 58.6 (3.4$\times$) & 75.0 (3.8$\times$) \\
5  & 1 & 90.5 (2.7$\times$) & 60.6 (2.0$\times$) & 75.6 (2.3$\times$) & 89.5 (3.3$\times$) & 53.2 (2.7$\times$) & 71.4 (3.0$\times$) & 87.0 (3.3$\times$) & 49.6 (2.8$\times$) & 68.3 (3.0$\times$) & 84.5 (3.1$\times$) & 46.8 (3.0$\times$) & 65.6 (3.0$\times$) \\
\bottomrule
\end{tabular}%
}
\end{table*}

\begin{table*}[ht]
\centering
\small
\renewcommand{\arraystretch}{1.12}
\caption{Ablation of score-threshold policies with \textbf{margin}-based token acceptance estimator. Static score uses \(s=\hat K-c\); dynamic score uses \(s=\hat K-c\cdot N_{\mathrm{hi}}\).}
\label{tab:ablation_policy_sdar_score_margin}
\resizebox{\linewidth}{!}{%
\begin{tabular}{cc ccc ccc ccc ccc}
\toprule
\(\tau_{\mathrm{score}}\) & \(c\) & \multicolumn{3}{c}{\textbf{B=4}} & \multicolumn{3}{c}{\textbf{B=8}} & \multicolumn{3}{c}{\textbf{B=16}} & \multicolumn{3}{c}{\textbf{B=32}} \\
\cmidrule(lr){3-5}\cmidrule(lr){6-8}\cmidrule(lr){9-11}\cmidrule(lr){12-14}
& & GSM8K & MBPP & Avg & GSM8K & MBPP & Avg & GSM8K & MBPP & Avg & GSM8K & MBPP & Avg \\
\midrule
\multicolumn{14}{c}{\textbf{Static score} \(s=\hat K-c\)} \\
\midrule
\multirow{3}{*}{-5}
& 1 & 93.0 (2.2$\times$) & 59.6 (2.1$\times$) & 76.3 (2.1$\times$) & 90.5 (3.1$\times$) & 56.2 (2.7$\times$) & 73.4 (2.9$\times$) & 89.5 (3.4$\times$) & 57.0 (3.1$\times$) & 73.2 (3.3$\times$) & 89.5 (3.7$\times$) & 57.2 (\textbf{3.6}$\times$) & 73.4 (3.6$\times$) \\
& 2 & 93.0 (2.2$\times$) & 59.6 (1.9$\times$) & 76.3 (2.1$\times$) & 90.5 (3.1$\times$) & 56.2 (2.7$\times$) & 73.4 (2.9$\times$) & 90.5 (3.5$\times$) & 57.0 (3.2$\times$) & 73.8 (3.3$\times$) & 89.0 (3.7$\times$) & 55.6 (3.2$\times$) & 72.3 (3.4$\times$) \\
& 4 & 93.0 (2.2$\times$) & \textbf{62.0} (2.0$\times$) & \textbf{77.5} (2.1$\times$) & 89.0 (3.1$\times$) & 56.6 (2.9$\times$) & 72.8 (3.0$\times$) & 89.5 (3.6$\times$) & 55.2 (3.2$\times$) & 72.4 (3.4$\times$) & 89.5 (3.7$\times$) & 55.6 (3.3$\times$) & 72.5 (3.5$\times$) \\
\cmidrule(lr){1-14}
\multirow{3}{*}{-1}
& 1 & 93.0 (2.2$\times$) & 61.8 (2.0$\times$) & 77.4 (2.1$\times$) & 91.0 (3.1$\times$) & \textbf{62.6} (2.4$\times$) & \textbf{76.8} (2.8$\times$) & 90.5 (4.1$\times$) & 61.4 (3.1$\times$) & \textbf{76.0} (3.6$\times$) & \textbf{92.5} (4.3$\times$) & 59.8 (3.1$\times$) & \textbf{76.2} (3.6$\times$) \\
& 2 & 88.0 (2.3$\times$) & 57.8 (1.9$\times$) & 72.9 (2.0$\times$) & 92.5 (3.1$\times$) & 56.0 (2.2$\times$) & 74.2 (2.6$\times$) & 89.5 (4.0$\times$) & 58.2 (3.0$\times$) & 73.8 (3.5$\times$) & 92.0 (4.0$\times$) & 55.6 (3.0$\times$) & 73.8 (3.4$\times$) \\
& 4 & 92.0 (2.2$\times$) & 61.2 (2.1$\times$) & 76.6 (2.1$\times$) & 89.5 (3.3$\times$) & 61.0 (2.6$\times$) & 75.2 (2.9$\times$) & 89.0 (\textbf{4.2}$\times$) & 59.2 (3.0$\times$) & 74.1 (3.5$\times$) & 88.0 (4.4$\times$) & \textbf{61.0} (3.2$\times$) & 74.5 (3.8$\times$) \\
\cmidrule(lr){1-14}
\multirow{3}{*}{0}
& 1 & 89.5 (2.1$\times$) & 59.8 (2.0$\times$) & 74.6 (2.1$\times$) & 89.0 (3.3$\times$) & 54.4 (2.8$\times$) & 71.7 (3.0$\times$) & 90.0 (3.5$\times$) & 54.4 (3.0$\times$) & 72.2 (3.2$\times$) & 85.5 (3.6$\times$) & 55.0 (3.4$\times$) & 70.2 (3.5$\times$) \\
& 2 & 90.5 (2.3$\times$) & 59.2 (2.1$\times$) & 74.8 (2.2$\times$) & 90.5 (3.4$\times$) & 55.6 (2.9$\times$) & 73.0 (3.2$\times$) & 86.0 (3.7$\times$) & 55.8 (3.1$\times$) & 70.9 (3.4$\times$) & 85.5 (3.6$\times$) & 52.8 (3.4$\times$) & 69.2 (3.5$\times$) \\
& 4 & 92.0 (2.3$\times$) & 60.0 (2.2$\times$) & 76.0 (2.2$\times$) & 93.0 (3.5$\times$) & 57.0 (\textbf{3.2}$\times$) & 75.0 (3.3$\times$) & 88.0 (3.7$\times$) & 54.8 (3.2$\times$) & 71.4 (3.5$\times$) & 87.5 (3.7$\times$) & 50.4 (3.5$\times$) & 69.0 (3.6$\times$) \\
\cmidrule(lr){1-14}
\multirow{3}{*}{1}
& 1 & 90.0 (2.3$\times$) & 59.2 (\textbf{2.3}$\times$) & 74.6 (2.3$\times$) & 89.5 (3.2$\times$) & 54.4 (2.5$\times$) & 72.0 (2.8$\times$) & 88.0 (3.9$\times$) & 57.6 (3.3$\times$) & 72.8 (3.6$\times$) & 85.0 (3.3$\times$) & 53.0 (3.3$\times$) & 69.0 (3.3$\times$) \\
& 2 & \textbf{93.5} (2.4$\times$) & 60.2 (2.2$\times$) & 76.8 (2.3$\times$) & 93.0 (3.3$\times$) & 56.2 (2.8$\times$) & 74.6 (3.0$\times$) & 87.0 (4.1$\times$) & 54.2 (3.3$\times$) & 70.6 (\textbf{3.7}$\times$) & 88.0 (3.3$\times$) & 57.0 (3.2$\times$) & 72.5 (3.2$\times$) \\
& 4 & 91.0 (\textbf{3.0}$\times$) & 60.6 (\textbf{2.3}$\times$) & 75.8 (\textbf{2.6}$\times$) & 90.5 (3.3$\times$) & 55.6 (2.8$\times$) & 73.0 (3.0$\times$) & 88.5 (3.7$\times$) & 54.2 (\textbf{3.5}$\times$) & 71.4 (3.6$\times$) & 86.0 (3.5$\times$) & 49.6 (3.1$\times$) & 67.8 (3.3$\times$) \\
\cmidrule(lr){1-14}
\multirow{3}{*}{5}
& 1 & 90.5 (2.7$\times$) & 61.4 (2.0$\times$) & 76.0 (2.3$\times$) & \textbf{93.5} (3.5$\times$) & 57.2 (3.0$\times$) & 75.4 (\textbf{3.3}$\times$) & 89.5 (3.5$\times$) & 50.6 (2.9$\times$) & 70.0 (3.2$\times$) & 84.5 (3.5$\times$) & 51.2 (3.2$\times$) & 67.8 (3.4$\times$) \\
& 2 & 91.0 (2.5$\times$) & 61.4 (2.0$\times$) & 76.2 (2.2$\times$) & 90.5 (\textbf{3.6}$\times$) & 56.2 (3.0$\times$) & 73.4 (\textbf{3.3}$\times$) & 86.5 (3.3$\times$) & 53.2 (2.9$\times$) & 69.8 (3.1$\times$) & 86.0 (3.3$\times$) & 48.8 (2.9$\times$) & 67.4 (3.1$\times$) \\
& 4 & 90.5 (2.7$\times$) & 61.4 (2.1$\times$) & 76.0 (2.4$\times$) & 89.5 (3.4$\times$) & 53.4 (2.8$\times$) & 71.5 (3.1$\times$) & 89.5 (3.5$\times$) & 51.6 (2.8$\times$) & 70.5 (3.1$\times$) & 86.0 (3.4$\times$) & 50.4 (3.0$\times$) & 68.2 (3.2$\times$) \\
\midrule
\multicolumn{14}{c}{\textbf{Dynamic score} \(s=\hat K-c\cdot N_{\mathrm{hi}}\)} \\
\midrule
-5 & 1 & 90.0 (1.9$\times$) & \textbf{62.0} (1.9$\times$) & 76.0 (1.9$\times$) & 90.0 (3.2$\times$) & 62.2 (2.7$\times$) & 76.1 (2.9$\times$) & 90.0 (3.9$\times$) & \textbf{61.8} (2.6$\times$) & \textbf{75.9} (3.1$\times$) & 88.5 (\textbf{4.6}$\times$) & \textbf{61.0} (3.3$\times$) & \textbf{74.8} (\textbf{3.9}$\times$) \\
-1 & 1 & 92.5 (2.1$\times$) & 60.2 (2.0$\times$) & \textbf{76.4} (2.0$\times$) & \textbf{93.5} (2.8$\times$) & 61.4 (2.3$\times$) & \textbf{77.5} (2.6$\times$) & 89.0 (4.0$\times$) & 59.6 (3.0$\times$) & 74.3 (3.5$\times$) & 90.5 (4.1$\times$) & 58.8 (2.9$\times$) & 74.6 (3.4$\times$) \\
0  & 1 & 93.0 (2.1$\times$) & 57.6 (1.9$\times$) & 75.3 (2.0$\times$) & 86.5 (3.3$\times$) & 59.0 (2.9$\times$) & 72.8 (3.1$\times$) & 88.5 (3.7$\times$) & 56.8 (3.1$\times$) & 72.6 (3.4$\times$) & 88.5 (3.5$\times$) & 56.0 (\textbf{3.6}$\times$) & 72.2 (3.6$\times$) \\
1  & 1 & \textbf{93.5} (\textbf{2.9}$\times$) & 58.6 (\textbf{2.3}$\times$) & 76.0 (\textbf{2.6}$\times$) & 90.0 (3.3$\times$) & 56.2 (2.8$\times$) & 73.1 (3.1$\times$) & \textbf{92.0} (3.7$\times$) & 54.8 (\textbf{3.3}$\times$) & 73.4 (\textbf{3.7}$\times$) & 87.0 (3.6$\times$) & 52.2 (3.3$\times$) & 69.6 (3.4$\times$) \\
5  & 1 & 90.5 (2.6$\times$) & 60.6 (2.0$\times$) & 75.6 (2.3$\times$) & 92.0 (\textbf{3.6}$\times$) & 55.0 (\textbf{3.0}$\times$) & 73.5 (\textbf{3.3}$\times$) & 85.5 (3.6$\times$) & 51.8 (3.1$\times$) & 68.7 (3.3$\times$) & 85.0 (3.4$\times$) & 51.2 (3.3$\times$) & 68.1 (3.3$\times$) \\
\bottomrule
\end{tabular}%
}
\end{table*}

\begin{table*}[ht]
\centering
\small
\renewcommand{\arraystretch}{1.12}
\caption{Ablation of hysteresis policies with \textbf{entropy}-based token acceptance estimator. We use dynamic score only, with fixed \(c=1\).}
\label{tab:ablation_policy_sdar_hysteresis}
\resizebox{\linewidth}{!}{%
\begin{tabular}{cc ccc ccc ccc ccc}
\toprule
\(\tau_{\mathrm{on}}\) & \(\tau_{\mathrm{off}}\) & \multicolumn{3}{c}{\textbf{B=4}} & \multicolumn{3}{c}{\textbf{B=8}} & \multicolumn{3}{c}{\textbf{B=16}} & \multicolumn{3}{c}{\textbf{B=32}} \\
\cmidrule(lr){3-5}\cmidrule(lr){6-8}\cmidrule(lr){9-11}\cmidrule(lr){12-14}
& & GSM8K & MBPP & Avg & GSM8K & MBPP & Avg & GSM8K & MBPP & Avg & GSM8K & MBPP & Avg \\
\midrule
\multirow{2}{*}{0}
& -5 & 90.5 (1.8$\times$) & \textbf{62.4} (1.9$\times$) & 76.4 (1.8$\times$) & \textbf{91.5} (3.2$\times$) & \textbf{63.6} (2.7$\times$) & \textbf{77.6} (3.0$\times$) & 89.0 (4.0$\times$) & 60.2 (2.6$\times$) & 74.6 (3.1$\times$) & 91.0 (4.3$\times$) & 59.2 (3.0$\times$) & 75.1 (3.5$\times$) \\
& -1 & \textbf{93.0} (2.1$\times$) & \textbf{62.4} (1.9$\times$) & \textbf{77.7} (2.0$\times$) & 89.5 (3.2$\times$) & \textbf{63.6} (2.6$\times$) & 76.6 (2.9$\times$) & 91.5 (\textbf{4.1}$\times$) & 60.2 (3.0$\times$) & 75.8 (3.5$\times$) & 91.0 (4.0$\times$) & \textbf{59.4} (3.1$\times$) & \textbf{75.2} (3.5$\times$) \\
\cmidrule(lr){1-14}
\multirow{2}{*}{1}
& -5 & 91.5 (2.1$\times$) & \textbf{62.4} (\textbf{2.0}$\times$) & 77.0 (2.0$\times$) & 89.0 (3.3$\times$) & 59.8 (2.7$\times$) & 74.4 (3.0$\times$) & 91.5 (3.8$\times$) & 59.4 (2.9$\times$) & 75.4 (3.3$\times$) & 90.5 (4.2$\times$) & 55.8 (\textbf{3.2}$\times$) & 73.2 (\textbf{3.7}$\times$) \\
& -1 & 91.5 (1.9$\times$) & \textbf{62.4} (1.9$\times$) & 77.0 (1.9$\times$) & 88.5 (3.1$\times$) & 59.8 (2.6$\times$) & 74.2 (2.9$\times$) & 91.0 (\textbf{4.1}$\times$) & \textbf{61.4} (\textbf{3.2}$\times$) & \textbf{76.2} (\textbf{3.6}$\times$) & \textbf{91.5} (\textbf{4.4}$\times$) & 57.4 (3.1$\times$) & 74.4 (3.6$\times$) \\
\cmidrule(lr){1-14}
\multirow{2}{*}{5}
& -5 & 91.0 (\textbf{2.5}$\times$) & 60.6 (\textbf{2.0}$\times$) & 75.8 (\textbf{2.3}$\times$) & 89.5 (\textbf{3.4}$\times$) & 53.6 (\textbf{2.8}$\times$) & 71.5 (\textbf{3.1}$\times$) & 92.5 (3.7$\times$) & 50.4 (2.9$\times$) & 71.5 (3.3$\times$) & 87.5 (3.7$\times$) & 51.6 (3.1$\times$) & 69.5 (3.4$\times$) \\
& -1 & 90.5 (2.4$\times$) & 60.6 (\textbf{2.0}$\times$) & 75.8 (\textbf{2.3}$\times$) & 91.0 (3.3$\times$) & 55.4 (\textbf{2.8}$\times$) & 73.2 (3.0$\times$) & \textbf{93.0} (3.3$\times$) & 51.6 (2.8$\times$) & 72.3 (3.1$\times$) & 88.5 (3.8$\times$) & 50.2 (3.0$\times$) & 69.3 (3.4$\times$) \\
\bottomrule
\end{tabular}%
}
\end{table*}

\begin{table*}[ht]
\centering
\small
\renewcommand{\arraystretch}{1.12}
\caption{Ablation of contextual-bandit UCB policies. The context tuple \((a,b,c)\) denotes the numbers of bins for mask span length, decoding progress, and entropy, respectively. We report results for different UCB exploration coefficients \(\beta\).}
\label{tab:ablation_policy_sdar_ucb}
\resizebox{\linewidth}{!}{%
\begin{tabular}{c ccc ccc ccc ccc}
\toprule
\(\beta\) & \multicolumn{3}{c}{\textbf{B=4}} & \multicolumn{3}{c}{\textbf{B=8}} & \multicolumn{3}{c}{\textbf{B=16}} & \multicolumn{3}{c}{\textbf{B=32}} \\
\cmidrule(lr){2-4}\cmidrule(lr){5-7}\cmidrule(lr){8-10}\cmidrule(lr){11-13}
& GSM8K & MBPP & Avg & GSM8K & MBPP & Avg & GSM8K & MBPP & Avg & GSM8K & MBPP & Avg \\
\midrule
\multicolumn{13}{c}{\textbf{Tuple} \((1,2,2)\)} \\
\midrule
0.2 & \textbf{92.5} (2.2$\times$) & 53.4 (1.9$\times$) & 73.0 (2.1$\times$) & 87.0 (3.4$\times$) & 49.0 (2.5$\times$) & 68.0 (2.9$\times$) & 87.0 (3.4$\times$) & 50.2 (2.7$\times$) & 68.6 (3.0$\times$) & 83.5 (4.2$\times$) & 52.6 (\textbf{3.2}$\times$) & 68.0 (3.6$\times$) \\
0.5 & 91.5 (2.2$\times$) & 55.4 (2.0$\times$) & 73.5 (2.1$\times$) & \textbf{89.5} (\textbf{3.5}$\times$) & 51.6 (\textbf{2.7}$\times$) & 70.5 (\textbf{3.1}$\times$) & 89.5 (3.6$\times$) & 52.0 (2.8$\times$) & \textbf{70.8} (3.2$\times$) & 82.5 (3.8$\times$) & 49.6 (2.8$\times$) & 66.0 (3.2$\times$) \\
1   & \textbf{92.5} (2.3$\times$) & 55.0 (2.0$\times$) & 73.8 (2.2$\times$) & 85.5 (3.4$\times$) & 47.6 (\textbf{2.7}$\times$) & 66.5 (3.0$\times$) & 86.0 (3.9$\times$) & 50.8 (2.9$\times$) & 68.4 (3.3$\times$) & 81.5 (3.5$\times$) & 48.6 (2.9$\times$) & 65.0 (3.2$\times$) \\
2   & 91.5 (2.3$\times$) & 53.6 (1.8$\times$) & 72.5 (2.0$\times$) & 87.5 (2.7$\times$) & 49.8 (2.4$\times$) & 68.7 (2.5$\times$) & 87.5 (3.8$\times$) & 48.0 (2.9$\times$) & 67.8 (3.3$\times$) & 83.0 (3.8$\times$) & 50.0 (3.0$\times$) & 66.5 (3.3$\times$) \\
5   & 89.5 (2.3$\times$) & 46.0 (2.0$\times$) & 67.8 (2.1$\times$) & 84.0 (2.8$\times$) & 44.6 (2.3$\times$) & 64.3 (2.5$\times$) & 85.0 (3.5$\times$) & 47.4 (2.7$\times$) & 66.2 (3.1$\times$) & 82.5 (3.7$\times$) & 46.8 (2.9$\times$) & 64.6 (3.3$\times$) \\
\midrule
\multicolumn{13}{c}{\textbf{Tuple} \((2,1,2)\)} \\
\midrule
0.2 & 90.0 (\textbf{2.5}$\times$) & 57.0 (2.1$\times$) & 73.5 (\textbf{2.3}$\times$) & 85.0 (3.1$\times$) & 55.2 (2.6$\times$) & 70.1 (2.8$\times$) & 85.0 (3.9$\times$) & 53.2 (\textbf{3.1}$\times$) & 69.1 (3.4$\times$) & 83.5 (3.8$\times$) & 52.8 (3.0$\times$) & 68.2 (3.4$\times$) \\
0.5 & 90.5 (\textbf{2.5}$\times$) & \textbf{61.4} (2.1$\times$) & \textbf{76.0} (\textbf{2.3}$\times$) & 88.5 (3.1$\times$) & 55.0 (2.6$\times$) & \textbf{71.8} (2.9$\times$) & 88.0 (\textbf{4.0}$\times$) & 50.2 (\textbf{3.1}$\times$) & 69.1 (\textbf{3.5}$\times$) & 85.5 (4.0$\times$) & 48.6 (2.6$\times$) & 67.0 (3.2$\times$) \\
1   & 89.5 (2.4$\times$) & 52.2 (2.0$\times$) & 70.9 (2.2$\times$) & 84.0 (2.9$\times$) & 53.8 (2.5$\times$) & 68.9 (2.7$\times$) & 84.5 (3.8$\times$) & 52.2 (3.0$\times$) & 68.3 (3.4$\times$) & 82.5 (3.6$\times$) & 50.2 (2.8$\times$) & 66.3 (3.2$\times$) \\
2   & 90.0 (2.2$\times$) & 50.4 (2.0$\times$) & 70.2 (2.1$\times$) & 86.0 (2.9$\times$) & 50.2 (2.6$\times$) & 68.1 (2.8$\times$) & 88.0 (3.7$\times$) & 50.6 (2.8$\times$) & 69.3 (3.2$\times$) & 81.5 (4.0$\times$) & 48.6 (2.9$\times$) & 65.0 (3.4$\times$) \\
5   & 88.5 (\textbf{2.5}$\times$) & 52.4 (2.0$\times$) & 70.5 (2.2$\times$) & 83.0 (2.9$\times$) & 45.2 (2.3$\times$) & 64.1 (2.6$\times$) & 87.0 (3.7$\times$) & 46.6 (2.9$\times$) & 66.8 (3.2$\times$) & 86.0 (4.0$\times$) & 46.8 (2.9$\times$) & 66.4 (3.4$\times$) \\
\midrule
\multicolumn{13}{c}{\textbf{Tuple} \((2,2,1)\)} \\
\midrule
0.2 & 91.0 (\textbf{2.5}$\times$) & 56.6 (\textbf{2.2}$\times$) & 73.8 (\textbf{2.3}$\times$) & 87.5 (3.2$\times$) & \textbf{55.6} (2.6$\times$) & 71.5 (2.9$\times$) & 84.5 (3.5$\times$) & 52.8 (2.7$\times$) & 68.7 (3.1$\times$) & \textbf{87.5} (4.1$\times$) & 49.8 (3.0$\times$) & 68.7 (3.5$\times$) \\
0.5 & 90.0 (2.2$\times$) & 57.2 (\textbf{2.2}$\times$) & 73.6 (2.2$\times$) & 89.0 (3.2$\times$) & 53.4 (\textbf{2.7}$\times$) & 71.2 (2.9$\times$) & 85.5 (3.3$\times$) & 53.2 (2.8$\times$) & 69.3 (3.0$\times$) & 85.0 (\textbf{4.4}$\times$) & 51.6 (3.0$\times$) & 68.3 (3.6$\times$) \\
1   & 91.0 (2.2$\times$) & 53.2 (2.0$\times$) & 72.1 (2.1$\times$) & 85.0 (3.3$\times$) & 51.8 (\textbf{2.7}$\times$) & 68.4 (3.0$\times$) & 84.5 (3.6$\times$) & 51.4 (2.7$\times$) & 68.0 (3.1$\times$) & 85.5 (4.2$\times$) & 50.0 (3.1$\times$) & 67.8 (3.6$\times$) \\
2   & 88.5 (2.1$\times$) & 53.0 (1.9$\times$) & 70.8 (2.0$\times$) & 85.0 (3.2$\times$) & 49.2 (2.6$\times$) & 67.1 (2.9$\times$) & 84.5 (3.5$\times$) & 50.2 (2.8$\times$) & 67.3 (3.1$\times$) & 86.5 (4.1$\times$) & 51.6 (2.9$\times$) & 69.0 (3.4$\times$) \\
5   & 90.0 (2.1$\times$) & 51.8 (1.8$\times$) & 70.9 (1.9$\times$) & 87.0 (3.1$\times$) & 45.8 (2.6$\times$) & 66.4 (2.8$\times$) & 84.5 (3.6$\times$) & 47.8 (2.8$\times$) & 66.1 (3.2$\times$) & 82.5 (3.9$\times$) & 46.0 (3.0$\times$) & 64.2 (3.4$\times$) \\
\midrule
\multicolumn{13}{c}{\textbf{Tuple} \((2,2,2)\)} \\
\midrule
0.2 & 91.0 (\textbf{2.5}$\times$) & 56.2 (2.1$\times$) & 73.6 (\textbf{2.3}$\times$) & 87.5 (3.0$\times$) & 54.8 (2.4$\times$) & 71.2 (2.7$\times$) & 84.0 (3.6$\times$) & 51.4 (3.0$\times$) & 67.7 (3.3$\times$) & 84.5 (3.9$\times$) & 49.8 (2.9$\times$) & 67.2 (3.3$\times$) \\
0.5 & \textbf{92.5} (2.4$\times$) & 57.2 (2.1$\times$) & 74.8 (2.2$\times$) & 87.5 (3.0$\times$) & 53.4 (2.5$\times$) & 70.5 (2.7$\times$) & 87.5 (3.6$\times$) & 53.2 (\textbf{3.1}$\times$) & 70.3 (3.4$\times$) & 83.5 (3.8$\times$) & 53.8 (2.9$\times$) & 68.7 (3.3$\times$) \\
1   & 91.5 (2.5$\times$) & 53.2 (\textbf{2.2}$\times$) & 72.4 (\textbf{2.3}$\times$) & 86.0 (3.1$\times$) & 51.0 (2.5$\times$) & 68.5 (2.8$\times$) & 86.0 (3.9$\times$) & 51.4 (3.0$\times$) & 68.7 (3.4$\times$) & 86.5 (3.8$\times$) & 50.0 (2.7$\times$) & 68.2 (3.2$\times$) \\
2   & 89.5 (2.4$\times$) & 50.6 (2.0$\times$) & 70.0 (2.2$\times$) & 84.5 (3.1$\times$) & 52.4 (2.2$\times$) & 68.5 (2.6$\times$) & 89.5 (3.6$\times$) & 51.4 (2.9$\times$) & 70.5 (3.2$\times$) & 86.5 (4.1$\times$) & 51.6 (2.9$\times$) & 69.0 (3.4$\times$) \\
5   & 90.0 (2.2$\times$) & 51.8 (2.0$\times$) & 70.9 (2.1$\times$) & 85.0 (3.0$\times$) & 45.8 (2.4$\times$) & 65.4 (2.7$\times$) & 84.0 (3.4$\times$) & 47.8 (2.6$\times$) & 65.9 (2.9$\times$) & 81.5 (3.9$\times$) & 46.0 (3.1$\times$) & 63.7 (3.4$\times$) \\
\midrule
\multicolumn{13}{c}{\textbf{Tuple} \((4,4,4)\)} \\
\midrule
0.2 & 88.0 (2.3$\times$) & 53.4 (2.0$\times$) & 70.7 (2.1$\times$) & \textbf{89.5} (3.3$\times$) & 51.8 (2.6$\times$) & 70.7 (2.9$\times$) & 87.5 (3.6$\times$) & 53.4 (2.8$\times$) & 70.5 (3.2$\times$) & 85.5 (4.0$\times$) & 51.8 (3.1$\times$) & 68.7 (3.5$\times$) \\
0.5 & 88.5 (2.1$\times$) & 54.8 (1.9$\times$) & 71.7 (2.0$\times$) & 86.0 (3.2$\times$) & 52.0 (2.6$\times$) & 69.0 (2.9$\times$) & 85.0 (3.6$\times$) & 53.6 (2.8$\times$) & 69.3 (3.1$\times$) & 83.0 (4.2$\times$) & \textbf{54.6} (3.1$\times$) & 68.8 (3.6$\times$) \\
1   & 89.5 (2.2$\times$) & 52.4 (1.9$\times$) & 71.0 (2.1$\times$) & 86.0 (3.2$\times$) & 51.4 (2.5$\times$) & 68.7 (2.8$\times$) & 88.5 (3.5$\times$) & 52.8 (2.9$\times$) & 70.7 (3.1$\times$) & 87.0 (4.1$\times$) & 53.8 (3.1$\times$) & \textbf{70.4} (3.6$\times$) \\
2   & 88.5 (2.2$\times$) & 51.2 (1.9$\times$) & 69.8 (2.0$\times$) & 86.5 (3.1$\times$) & 49.0 (2.6$\times$) & 67.8 (2.8$\times$) & \textbf{90.0} (3.6$\times$) & \textbf{54.8} (2.8$\times$) & 68.4 (3.1$\times$) & 84.5 (4.3$\times$) & 52.6 (\textbf{3.2}$\times$) & 68.5 (\textbf{3.7}$\times$) \\
5   & 89.5 (2.1$\times$) & 51.0 (2.0$\times$) & 70.2 (2.0$\times$) & 83.0 (3.0$\times$) & 52.0 (2.5$\times$) & 67.5 (2.7$\times$) & 89.0 (3.6$\times$) & 52.8 (2.8$\times$) & 70.9 (3.2$\times$) & 82.5 (4.0$\times$) & 52.6 (3.0$\times$) & 67.5 (3.4$\times$) \\
\bottomrule
\end{tabular}%
}
\end{table*}

\begin{figure}[t]
    \centering

    \begin{subfigure}[t]{0.25\linewidth}
        \centering
        \includegraphics[width=\linewidth]{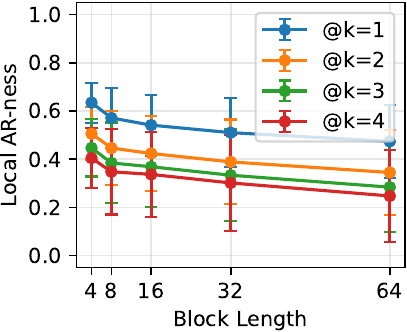}
        \caption{\small \texttt{GSM8K} local AR-ness}
        \label{fig:sdar_arness_local_gsm8k}
    \end{subfigure}%
    \begin{subfigure}[t]{0.25\linewidth}
        \centering
        \includegraphics[width=\linewidth]{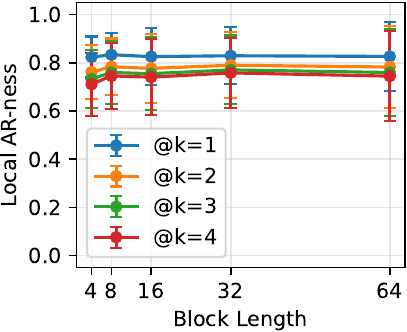}
        \caption{\small \texttt{MBPP} local AR-ness}
        \label{fig:sdar_arness_local_mbpp}
    \end{subfigure}%
    \begin{subfigure}[t]{0.25\linewidth}
        \centering
        \includegraphics[width=\linewidth]{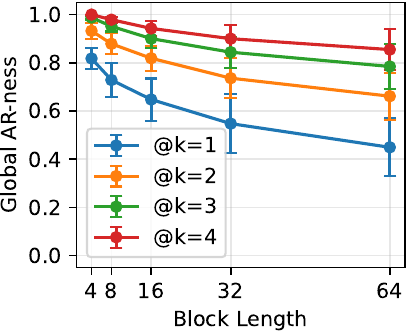}
        \caption{\small \texttt{GSM8K} global AR-ness}
        \label{fig:sdar_arness_global_gsm8k}
    \end{subfigure}%
    \begin{subfigure}[t]{0.25\linewidth}
        \centering
        \includegraphics[width=\linewidth]{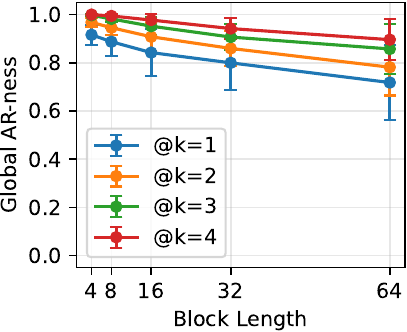}
        \caption{\small \texttt{MBPP} global AR-ness}
        \label{fig:sdar_arness_global_mbpp}
    \end{subfigure}

    \vspace{0.2em}

    \begin{subfigure}[t]{0.25\linewidth}
        \centering
        \includegraphics[width=\linewidth]{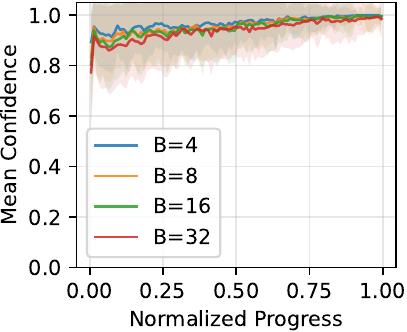}
        \caption{\small \texttt{GSM8K} static conf.}
        \label{fig:sdar_conf_static_gsm8k}
    \end{subfigure}%
    \begin{subfigure}[t]{0.25\linewidth}
        \centering
        \includegraphics[width=\linewidth]{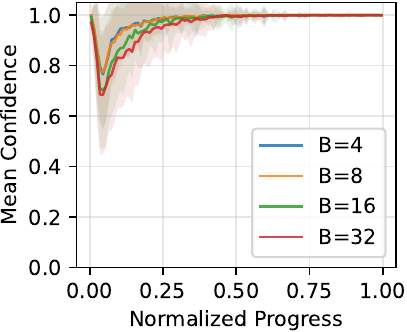}
        \caption{\small \texttt{MBPP} static conf.}
        \label{fig:sdar_conf_static_mbpp}
    \end{subfigure}%
    \begin{subfigure}[t]{0.25\linewidth}
        \centering
        \includegraphics[width=\linewidth]{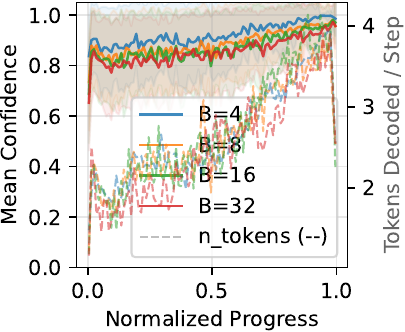}
        \caption{\small \texttt{GSM8K} dynamic conf.}
        \label{fig:sdar_conf_dynamic_gsm8k}
    \end{subfigure}%
    \begin{subfigure}[t]{0.25\linewidth}
        \centering
        \includegraphics[width=\linewidth]{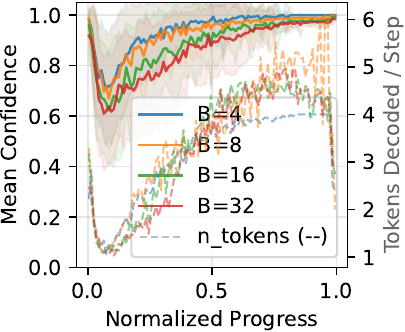}
        \caption{\small \texttt{MBPP} dynamic conf.}
        \label{fig:sdar_conf_dynamic_mbpp}
    \end{subfigure}

    \caption{AR-ness and decoding confidence statistics for SDAR-8B-Chat. Top row: local/global AR-ness on GSM8K and MBPP. Bottom row: normalized confidence statistics under static and dynamic decoding on GSM8K and MBPP.}
    \label{fig:sdar_arness_confidence}
\end{figure}

\begin{figure}[t]
    \centering

    \begin{subfigure}[t]{0.25\linewidth}
        \centering
        \includegraphics[width=\linewidth]{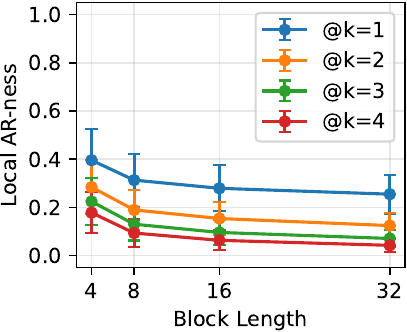}
        \caption{\small \texttt{GSM8K} local AR-ness}
        \label{fig:llada_arness_local_gsm8k}
    \end{subfigure}%
    \begin{subfigure}[t]{0.25\linewidth}
        \centering
        \includegraphics[width=\linewidth]{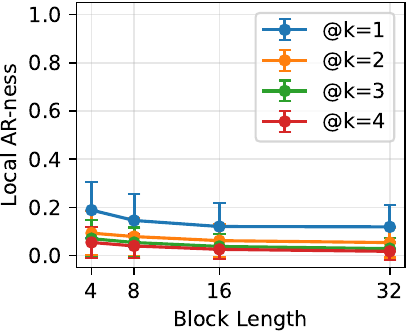}
        \caption{\small \texttt{MBPP} local AR-ness}
        \label{fig:llada_arness_local_mbpp}
    \end{subfigure}%
    \begin{subfigure}[t]{0.25\linewidth}
        \centering
        \includegraphics[width=\linewidth]{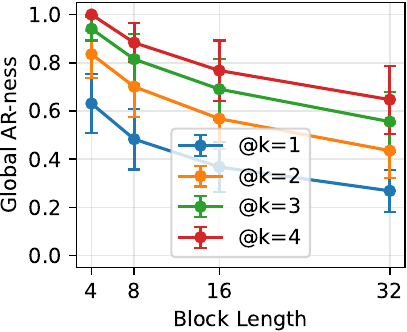}
        \caption{\small \texttt{GSM8K} global AR-ness}
        \label{fig:llada_arness_global_gsm8k}
    \end{subfigure}%
    \begin{subfigure}[t]{0.25\linewidth}
        \centering
        \includegraphics[width=\linewidth]{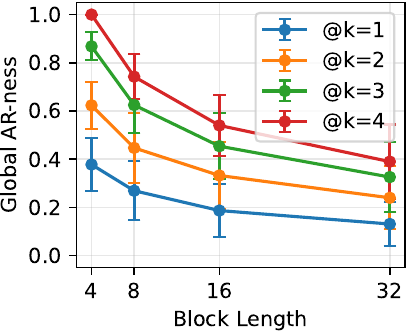}
        \caption{\small \texttt{MBPP} global AR-ness}
        \label{fig:llada_arness_global_mbpp}
    \end{subfigure}

    \vspace{0.2em}

    \begin{subfigure}[t]{0.25\linewidth}
        \centering
        \includegraphics[width=\linewidth]{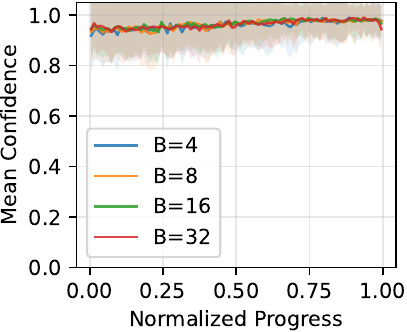}
        \caption{\small \texttt{GSM8K} static conf.}
        \label{fig:llada_conf_static_gsm8k}
    \end{subfigure}%
    \begin{subfigure}[t]{0.25\linewidth}
        \centering
        \includegraphics[width=\linewidth]{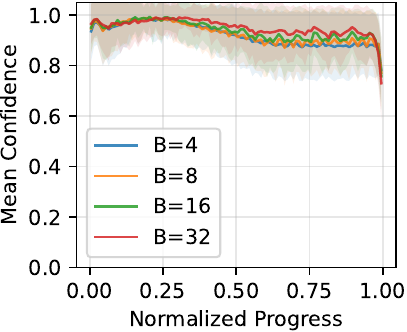}
        \caption{\small \texttt{MBPP} static conf.}
        \label{fig:llada_conf_static_mbpp}
    \end{subfigure}%
    \begin{subfigure}[t]{0.25\linewidth}
        \centering
        \includegraphics[width=\linewidth]{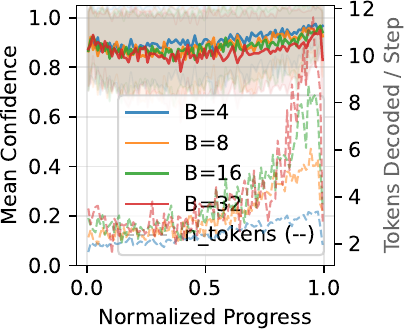}
        \caption{\small \texttt{GSM8K} dynamic conf.}
        \label{fig:llada_conf_dynamic_gsm8k}
    \end{subfigure}%
    \begin{subfigure}[t]{0.25\linewidth}
        \centering
        \includegraphics[width=\linewidth]{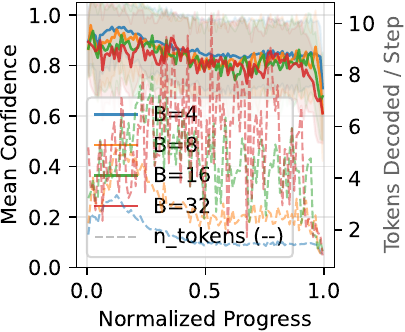}
        \caption{\small \texttt{MBPP} dynamic conf.}
        \label{fig:llada_conf_dynamic_mbpp}
    \end{subfigure}

    \caption{AR-ness and decoding confidence statistics for LLaDA-2.1-Mini. Top row: local/global AR-ness on GSM8K and MBPP. Bottom row: normalized confidence statistics under static and dynamic decoding on GSM8K and MBPP.}
    \label{fig:llada_arness_confidence}
\end{figure}

\subsection{Ablation on rejection-sampling ratio tempering}
\label{app:ratio_tempering}

\paragraph{Ratio tempering.}
Table~\ref{tab:ablation_ratio_tempering} studies tempering of the rejection-sampling ratio, replacing $q_i/p_i$ with $(q_i/p_i)^{\gamma}$. Smaller $\gamma$ makes acceptance more aggressive, while larger $\gamma$ makes it more conservative. We observe that a slightly larger value, e.g.\ $\gamma=1.25$, can give a small accuracy improvement in some settings, especially for config-B, but typically with a minor speed degradation. Overall, the default choice $\gamma=1$ already provides a good tradeoff and is used in the main experiments.

\begin{table}[t]
\centering
\small
\renewcommand{\arraystretch}{1.12}
\setlength{\tabcolsep}{5pt}
\caption{Ablation of rejection-sampling ratio tempering for SSD on SDAR-8B-Chat. We vary the tempering factor \(\gamma\) in the acceptance ratio \((q_i/p_i)^\gamma\) and report accuracy (speedup relative to \(\gamma{=}1\)).}
\label{tab:ablation_ratio_tempering}
\begin{tabular}{c ccc ccc}
\toprule
\multirow{2}{*}{\(\gamma\)} & \multicolumn{3}{c}{\textbf{Config-A}} & \multicolumn{3}{c}{\textbf{Config-B}} \\
\cmidrule(lr){2-4}\cmidrule(lr){5-7}
& GSM8K & MBPP & Avg & GSM8K & MBPP & Avg \\
\midrule
0.5  & 89.5 (1.02$\times$) & 60.0 (1.06$\times$) & 74.8 (1.03$\times$) & 88.5 (1.02$\times$) & 58.0 (1.00$\times$) & 73.2 (1.01$\times$) \\
0.75 & 89.5 (0.99$\times$) & 62.0 (1.06$\times$) & 75.8 (1.01$\times$) & \textbf{90.5} (0.99$\times$) & 58.5 (0.95$\times$) & 74.5 (0.98$\times$) \\
1.0  & 89.5 (1.00$\times$) & \textbf{64.0} (1.00$\times$) & \textbf{76.8} (1.00$\times$) & 89.5 (1.00$\times$) & \textbf{60.5} (1.00$\times$) & 75.0 (1.00$\times$) \\
1.25 & 89.5 (1.00$\times$) & 60.5 (1.07$\times$) & 75.0 (1.02$\times$) & \textbf{90.5} (0.93$\times$) & \textbf{60.5} (0.96$\times$) & \textbf{75.5} (0.94$\times$) \\
1.5  & \textbf{90.5} (1.00$\times$) & 60.5 (0.96$\times$) & 75.5 (0.99$\times$) & \textbf{90.5} (1.01$\times$) & 60.0 (0.96$\times$) & 75.2 (1.00$\times$) \\
\bottomrule
\end{tabular}
\end{table}

\end{document}